\documentclass[final]{cvpr}

\usepackage{times}
\usepackage{epsfig}
\usepackage{graphicx}
\usepackage{amsmath}
\usepackage{amssymb}
\usepackage{nicefrac}
\usepackage{mathtools}
\usepackage{physics}
\usepackage{gensymb}
\usepackage{booktabs}
\usepackage{tabularx}
\usepackage{enumitem}
\usepackage[dvipsnames,svgnames,x11names]{xcolor}

\DeclarePairedDelimiter{\ceil}{\lceil}{\rceil}

\usepackage[pagebackref=true,breaklinks=true,colorlinks,bookmarks=false]{hyperref}

\def\cvprPaperID{1668} 
\def\confYear{CVPR 2021}
\begin{document}

\title{Weakly-Supervised Physically Unconstrained Gaze Estimation}

\author{
Rakshit Kothari\textsuperscript{1,2}\thanks{Rakshit Kothari was an intern at NVIDIA during the project.} \hspace{2mm}
Shalini De Mello\textsuperscript{1} \hspace{2mm}
Umar Iqbal\textsuperscript{1}\\
Wonmin Byeon\textsuperscript{1} \hspace{2mm}
Seonwook Park\textsuperscript{3} \hspace{2mm}
Jan Kautz\textsuperscript{1} \\
\vspace{1mm}
\textsuperscript{1}NVIDIA \qquad \textsuperscript{2}Rochester Institute of Technology \qquad \textsuperscript{3}Lunit Inc.\\
{\tt\small rsk3900@rit.edu; spark@lunit.io} \\ 
\tt\small \{shalinig, uiqbal, wbyeon, jkautz\}@nvidia.com
}
\maketitle

\begin{abstract}
A major challenge for physically unconstrained gaze estimation is acquiring training data with 3D gaze annotations for in-the-wild and outdoor scenarios. In contrast, videos of human interactions in unconstrained environments are abundantly available and can be much more easily annotated with frame-level activity labels. In this work, we tackle the previously unexplored problem of weakly-supervised gaze estimation from videos of human interactions. We leverage the insight that strong gaze-related geometric constraints exist when people perform the activity of ``looking at each other” (LAEO). To acquire viable 3D gaze supervision from LAEO labels, we propose a training algorithm along with several novel loss functions especially designed for the task. With weak supervision from two large scale CMU-Panoptic and AVA-LAEO activity datasets, we show significant improvements in (a) the accuracy of semi-supervised gaze estimation and (b) cross-domain generalization on the state-of-the-art physically unconstrained in-the-wild Gaze360 gaze estimation benchmark. We open source our code at \href{https://github.com/NVlabs/weakly-supervised-gaze}{https://github.com/NVlabs/weakly-supervised-gaze}. 

\end{abstract}
\vspace{-3mm}
\section{Introduction}
\vspace{-2mm}
\begin{figure}
    \centering
    \includegraphics[width=0.48\textwidth]{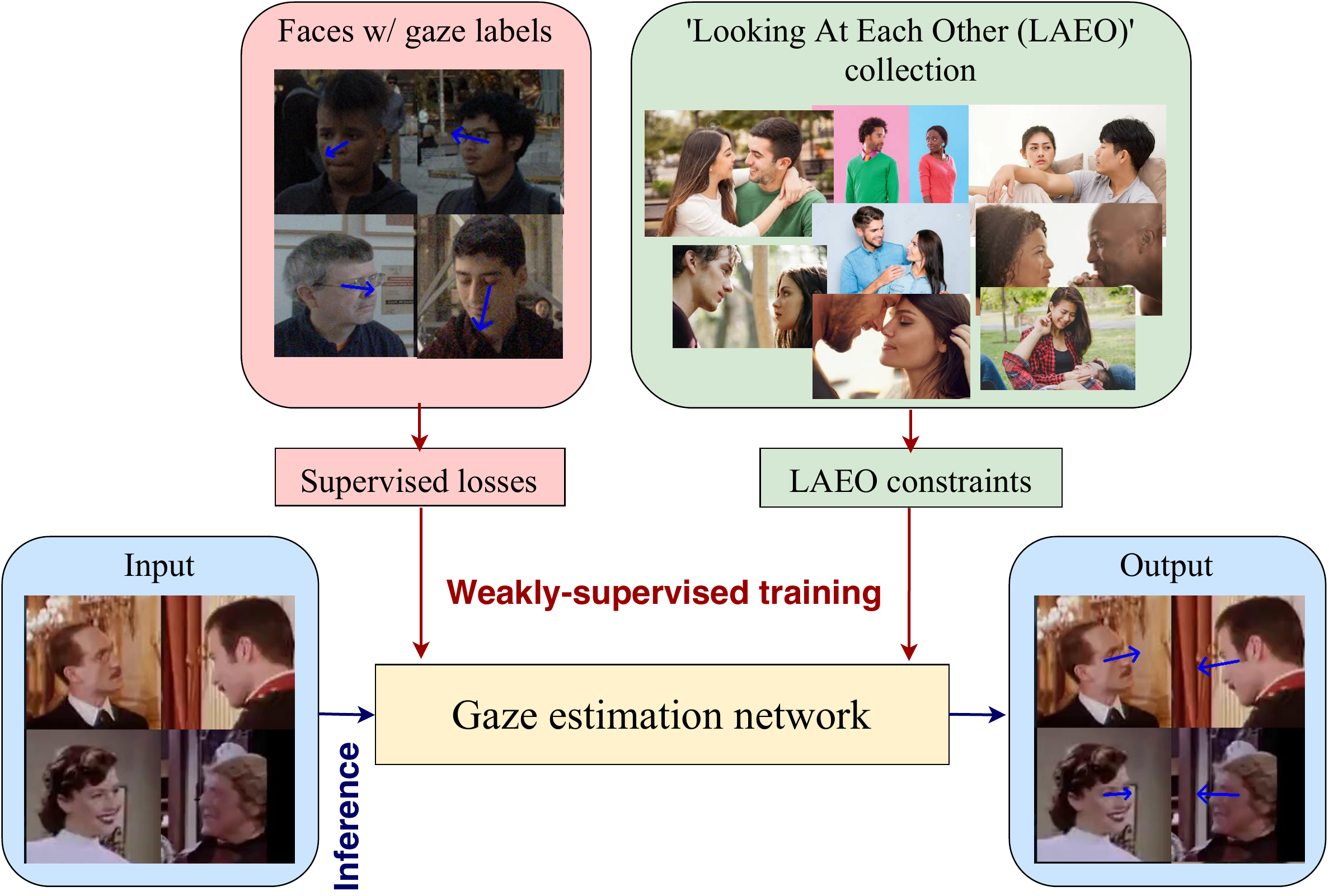}
    \caption{Overview of our weakly-supervised gaze estimation approach. We employ large collections of videos of people ``looking at each other" (LAEO) curated from the Internet without any explicit 3D gaze labels, either by themselves or in a semi-supervised manner to learn 3D gaze in physically unconstrained settings.
    }
    \label{fig:teaser}
    \vspace*{-5mm}
\end{figure}

Much progress has been made recently in the task of remote 3D gaze estimation from monocular images, but most of these methods are constrained to largely frontal subjects viewed by cameras located within a meter of them~\cite{Zhang2015CVPR,Krafka2016CVPR}. To go beyond frontal faces, a few recent works explore the more challenging problem of so-called ``physically unconstrained gaze estimation'', where larger camera-to-subject distances and higher variations in head pose and eye gaze angles are present~\cite{Kellnhofer2019, Zhang2020ETHXGaze,Fischer2018}. A significant challenge there is in acquiring training data with 3D gaze labels, generally and more so outdoors. Fortunately, several 3D gaze datasets with large camera-to-subject distances and variability in head pose have been collected recently in indoor laboratory environments using specialized multi-cameras setups~\cite{Yu2020HUMBI, Fischer2018, Zhang2020ETHXGaze, Park2020ECCV}. In contrast, the recent Gaze360 dataset~\cite{Kellnhofer2019} was collected both indoors and outdoors, at greater distances to subjects. While the approach of Gaze360 advances the field significantly, it nevertheless requires expensive hardware and many co-operative subjects and hence can be difficult to scale.

Recently ``weakly-supervised'' approaches have been demonstrated on various human perception tasks, such as body pose estimation via multi-view constraints~\cite{rhodin2018learning,iqbal2020weakly}, hand pose estimation via bio-mechanical constraints~\cite{spurr2020weakly}, and face reconstruction via differentiable rendering~\cite{deng2019accurate}. Nevertheless, little attention has been paid to exploring methods with weak supervision for frontal face gaze estimation~\cite{yu2020unsupervised} and none at all for physically unconstrained gaze estimation. Eye gaze is a natural and strong non-verbal form of human communication~\cite{mazur1980physiological}. For instance, babies detect and follow a caregiver's gaze from as early as four months of age~\cite{striano2006social}. Consequently, videos of human interactions involving eye gaze are commonplace and are abundantly available on the Internet~\cite{Gu2018}. 
Thus we pose the question: \emph{``Can machines learn to estimate 3D gaze by observing videos of humans interacting with each other?''}.

In this work, we tackle the previously unexplored problem of weakly supervising 3D gaze learning from videos of human interactions curated from the Internet (Fig.~\ref{fig:teaser}). We target the most challenging problem within this domain of physically unconstrained gaze estimation. Specifically, to learn 3D gaze we leverage the insight that strong gaze-related geometric constraints exist when people perform the commonplace interaction of ``looking at each other” (LAEO), \textit{i.e.}, the 3D gaze vectors of the two people interacting are oriented in opposite directions to each other. Videos of the LAEO activity can be easily curated from the Internet and annotated with frame-level labels for the presence of the LAEO activity and with 2D locations of the persons performing it~\cite{Marin-Jimenez2014,Marin-Jimenez2019}. However, estimating 3D gaze from just 2D LAEO annotations is challenging and ill-posed because of the depth ambiguity of the subjects in the scene. Furthermore, naively enforcing the geometric constraint of opposing gaze vector predictions for the two subjects performing LAEO is, by itself, insufficient supervision to avoid degenerate solutions while learning 3D gaze.

To solve these challenges and to extract viable 3D gaze supervision from weak LAEO labels, we propose a training algorithm that is especially designed for the task. We enforce several scene-level geometric 3D and 2D LAEO constraints between pairs of faces, which significantly aid in accurately learning 3D gaze information. While training, we also employ a self-training procedure and compute stronger pseudo 3D gaze labels from weak noisy estimates for pairs of faces in LAEO in an uncertainty-aware manner. Lastly, we employ an aleatoric gaze uncertainty loss and a symmetry loss to supervise learning. Our algorithm operates both in a purely weakly-supervised manner with LAEO data only or in a semi-supervised manner along with limited 3D gaze-labeled data.

We evaluate the real-world efficacy of our approach on the large physically unconstrained Gaze360~\cite{Kellnhofer2019} benchmark. We conduct various within- and cross-dataset experiments and obtain LAEO labels from two large-scale datasets: (a) the CMU Panoptic~\cite{Joo2019} with known 3D scene geometry and (b) the in-the-wild AVA-LAEO activity dataset~\cite{Marin-Jimenez2019} containing Internet videos. We show that our proposed approach can successfully learn 3D gaze information from weak LAEO labels. Furthermore, when combined with limited (in terms of the variability of subjects, head poses or environmental conditions) 3D gaze-labeled data in a semi-supervised setting, our approach can significantly help to improve accuracy and cross-domain generalization. Hence, our approach not only reduces the burden of acquiring data and labels for the task of physically unconstrained gaze estimation, but also helps to generalize better for diverse/naturalistic environments.

To summarize, our key contributions are:
\vspace{-3mm}
\begin{itemize}\setlength\itemsep{-1mm}
  \item We propose a novel weakly-supervised framework for learning 3D gaze from in-the-wild videos of people performing the activity of ``looking at each other". To our understanding, we are the first to employ videos of humans interacting to supervise 3D gaze learning.
  \item To effectively derive 3D gaze supervision from weak LAEO labels, we introduce several novel training objectives. We learn to predict aleatoric uncertainty, use it to derive strong pseudo-3D gaze labels, and further propose geometric LAEO 3D and 2D constraints to learn gaze from LAEO labels.
  \item Our experiments on the Gaze360 benchmark show that LAEO data can effectively augment data with strong 3D gaze labels both within and across datasets.
\end{itemize}

\section{Related Work}
\paragraph{3D Gaze Estimation} Recent developments in remote gaze estimation increasingly benefit from large-scale datasets with gaze direction~\cite{Zhang2015CVPR,FunesMora2014ETRA,Smith2013UIST,Fischer2018} or target~\cite{Krafka2016CVPR,Huang2017MVA} labels.
While earlier methods study the effect of different input facial regions~\cite{Krafka2016CVPR,Zhang2017CVPRW,Fischer2018,zhang20_bmvc}, later methods attempt to introduce domain-specific insights into their solutions. For example by encoding the eye-shape into the learning procedure~\cite{park2018deep,park2018learning,yu2020unsupervised,wang2018hierarchical}, or by considering the dependency between head orientation and gaze direction~\cite{zhu2017monocular,ranjan2018light,wang2019generalizing}, or modelling uncertainty or random effects~\cite{xiong2019mixed,cheng2018appearance,Kellnhofer2019}.
Other works propose few-shot adaptation approaches for improving performance for end-users~\cite{Park2019ICCV,linden2019learning,He_2019_ICCV,chen2020offset,liu2019differential}.
However, most such approaches restrict their evaluations to screen-based settings (with mostly frontal faces and subjects located within 1m of the camera) due to limitations in the diversity of available training datasets.

Recently proposed datasets such as RT-GENE~\cite{Fischer2018}, HUMBI~\cite{Yu2020HUMBI}, and ETH-XGaze~\cite{Zhang2020ETHXGaze} attempt to allow for gaze estimation in more physically unconstrained settings such as from profile faces of subjects located further from the camera.
As complex multi-view imaging setups are required, these datasets are inevitably collected in controlled laboratory conditions.
A notable exception is Gaze360~\cite{Kellnhofer2019}, which uses a panoramic camera for collecting data from multiple participants at once, both outdoors and indoors.
Yet, such collection methods are still difficult to scale compared to data sourced from the web, or via crowd-sourced participation such as done for the GazeCapture dataset~\cite{Krafka2016CVPR}.

In terms of learning a generalized gaze estimator using only small amounts of labeled data (without supervised pre-training), Yu~\etal~\cite{yu2020unsupervised} are the only prior art. However, their method is restricted to mostly frontal faces and assumes little to no movement of the head between pairs of samples from a given participant -- an assumption that does not hold in less constrained settings.

\vspace{-4mm}
\paragraph{Gaze Following and Social Interaction Labels} Given an image with a human, gaze following concerns the prediction of the human's gaze target position.
Performing this task with deep neural networks was initially explored by Recasens~\etal~\cite{recasens2015they}, with extensions to time sequence data and multiple camera views in~\cite{recasens2017following}.
Chong~\etal~\cite{chong2018connecting} improve performance on the static gaze following task further by jointly training to predict 3D gaze direction using the EYEDIAP dataset~\cite{FunesMora2014ETRA}, and by explicitly predicting whether the target is in frame. This work is also extended to video data in~\cite{Chong2020}. Much like the task of physically unconstrained gaze estimation, gaze following also involves viewing human subjects in all head poses, in diverse environments, and from larger distances. 
However, gaze following datasets are complex to annotate, and do not lend themselves well to the task of learning to predict 3D gaze due to the lack of scene and object geometry information. 

Alternatively, weak annotations for gaze-based interaction exist in the form of social interaction labels.
One such condition is the commonplace ``looking at each other" condition, also known as LAEO~\cite{Marin-Jimenez2014}, where a binary label is assigned to pairs of human heads for when they are gazing at each other. This is a simpler quantity to annotate compared to mutual attention or visual focus of attention. The recently published AVA-LAEO dataset~\cite{Marin-Jimenez2019} is an extension of the AVA dataset~\cite{Gu2018} and demonstrates the ease of acquiring such annotations for existing videos. To the best of our knowledge, we are the first to show that social interaction labels such as LAEO can be used for weakly-supervised gaze estimation. Furthermore, adding LAEO-based constraints and objectives consistently improves performance in cross-dataset and semi-supervised gaze estimation, further validating the real-world efficacy of our approach.
\section{Weakly-supervised Gaze Learning}

\begin{figure*}
    \centering
    \includegraphics[width=0.9\textwidth]{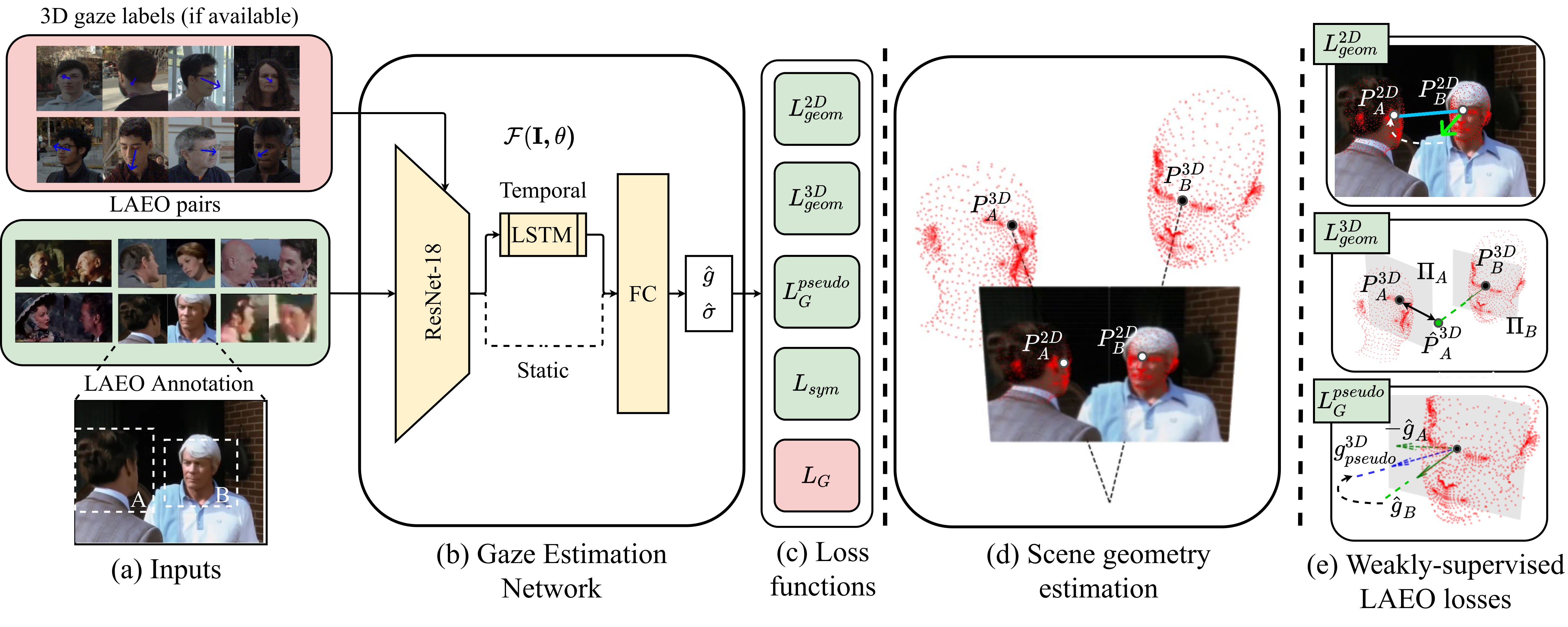}
    \caption{An overview of our weakly-supervised approach to learning 3D gaze from the ``looking at each other" human activity videos. From left to right, we show (a) the inputs to our gaze estimation network, \textit{i.e.}, pairs of head crops in LAEO with their scene images for weakly-supervised training and optionally single head crops with gaze labels for semi-supervised training (if available); (b) our gaze estimation network, which predicts gaze $\hat{g}$ and its uncertainty $\hat{\sigma}$; (c) the various weakly-supervised and fully-supervised losses used for training; (d) estimation of scene geometry for in-the-wild LAEO videos acquired from the web including that of the 2D and 3D positions of the cyclopean eyes of the subject pairs in LAEO used to compute the LAEO losses; and (e) details of our proposed LAEO losses including the geometric 2D LAEO loss ($L^{2D}_{geom}$), geometric 3D LAEO loss ($L^{3D}_{geom}$) and the pseudo gaze loss ($L^{pseudo}_G$).}
    \label{fig:summary_methods}
    \vspace{-3mm}
\end{figure*}

\subsection{Problem Definition and Motivation}
\vspace{-2mm}
Our goal is to supervise 3D gaze learning with weak supervision from in-the-wild videos of humans ``looking at each other". Such scenes contain the LAEO constraint, \textit{i.e.}, the 3D gazes of the two subjects are oriented along the same line, but in opposite directions to each other. We specifically target the challenging task of physically unconstrained gaze estimation where large subject-to-camera distances, and variations in head poses and environments are present. We assume that we have a large collection of videos containing LAEO activities available to us which can be acquired, for example, by searching the web with appropriate textual queries. We further assume that, by whatever means, the specific frames of a longer video sequence containing the LAEO activity have been located and that the 2D bounding boxes of the pair of faces in the LAEO condition are also available. We refer to these labels collectively as the ``LAEO labels".

Acquiring LAEO data is a relatively quick and cost effective way to curate lots of diverse training data. Nevertheless, Internet videos with LAEO labels cannot provide precise 3D gaze supervision. This is because, for such videos neither the scene's precise geometry, nor the camera's intrinsic parameters are known \textit{a priori}. Moreover, trivially enforcing the simple LAEO constraint of requiring the predicted gaze estimates of the two individuals to be opposite to each other is not sufficient for learning gaze. It quickly leads to degenerate solutions.

To address these various challenges, we design a novel weakly-supervised learning framework for 3D gaze estimation from LAEO data. Specifically, we propose a number of novel geometric scene-level LAEO losses, including a 3D and a 2D one, that are applied to pairs of faces in LAEO. For individual face inputs we also use an aleatoric gaze loss~\cite{Kendall2017}, which computes gaze uncertainty, along with a self-supervised symmetry loss. We further propose an uncertainty-aware self-training procedure to generate 3D gaze pseudo ground truth labels from pairs of faces exhibiting LAEO. Our training framework operates in two configurations: (a) a purely weakly-supervised one with LAEO data only and (b) a semi-supervised one, where LAEO data is combined with 3D gaze-labeled data.

\vspace{-1mm}
\subsection{Solution Overview} 
\vspace{-2mm}
Our overall framework for weakly-supervised 3D gaze learning from LAEO data is shown in Fig.~\ref{fig:summary_methods}. We wish to train the function $\mathcal{F(\bf{I}, \theta)}$ with weights $\theta$ to estimate gaze by providing video sequences of pairs of people exhibiting LAEO. Inspired by~\cite{Kellnhofer2019}, our gaze estimation network $\mathcal{F(\bf{I}, \theta)}$ consists of a ResNet-18 backbone followed by two bi-directional LSTM layers and a fully-connected (FC) layer, which estimates a gaze value $\hat{g} = \{\hat{g}_{\theta}, \hat{g}_{\phi}\}$ along with an uncertainty value $\hat{\sigma}$ corresponding to the central image in a sequence of 7 consecutive input frames. Here $\hat{g}_{\theta}$ and $\hat{g}_{\phi}$ indicate the estimates for the gaze pitch and yaw angles, respectively. In addition to this temporal version of our network, we also explore a \textit{static} variant, which takes a single image as input and bypasses the LSTM layers to directly connect the output of the backbone CNN to the FC layer.

For LAEO data, the input to our network is a pair of head crops of size $224\times224\times3$ each containing one of the two faces that exhibit LAEO along with the original scene image. No 3D gaze labels are available during training with LAEO data. If data containing explicit 3D gaze labels is additionally available for semi-supervised training, we extract single head crops from the scene images and input them along with their known ground truth 3D gaze labels into the network for training.

\vspace{-1mm}
\subsection{Loss Functions}
\label{sec:methods:laeo_geom}
\vspace{-2mm}
We employ several end-to-end differentiable geometric loss functions, which are derived from the LAEO constraint to supervised 3D gaze learning. These include two scene-level geometric 2D and 3D LAEO losses. We start by describing our technique for scene geometry estimation and 3D gaze origin determination for in-the-wild videos and then describe our geometric LAEO losses. We then describe our uncertainty-aware 3D gaze pseudo labeling procedure, followed by two additional losses -- the aleatoric gaze and symmetry losses that are applied to individual face inputs.

\vspace{-4mm}
\paragraph{Scene Geometry Estimation}
\label{sec:scene_geometry}
The geometric LAEO loss functions can only be computed in a coordinate system common to both subjects, \ie, the camera coordinate system. For Internet videos, we cannot reliably recover camera parameters or the subjects' 3D poses. So we instead derive approximate values for them. We approximate the camera focal parameter $f$ to be the size of the larger image dimension in pixels. The principal point is assumed to be at the center of the image. We detect the 2D facial landmarks of a subject using AlphaPose~\cite{fang2017rmpe} and refer to the midpoint of their left and right eye pixel locations as their ``2D cyclopean eye" $P^{2D} = (x,y)$. We assume it to be the point from where gaze originates for a subject on the 2D image plane. To find its 3D counterpart, \ie, the 3D cyclopean eye $P^{3D}$, we also estimate depth $z$ per subject and back-project $P^{2D}$ to 3D as $(zx/f, zy/f, z)$. This procedure ensures that $P^{2D}$ and $P^{3D}$ lie on the same projection line originating from the camera's center.

To recover depth $z$ of each subject, we first estimate their 2D-3D correspondences using DensePose~\cite{Guler2018}. We use the predicted 2D facial key-points~\cite{fang2017rmpe} and an average gender neutral 3D SMPL head model~\cite{SMPL:2015} to compute the 3D transformation required to fit the 3D head model to a particular subject using PnP~\cite{lepetit2009epnp}. This allows for an estimation of the 3D head model's location and orientation in the camera coordinate system, which in turn provides us with depth estimates in meters (see Fig.~\ref{fig:summary_methods}) for each subject. Specifically, we utilize the depth $z$ value of the mid points of the left and right eyes of the fitted 3D head model to recover depth of each subject. The end result is a shared 3D coordinate system for both subjects under LAEO (see Fig.~\ref{fig:summary_methods}). In  Sec. D.3 of the supplementary we further discuss the effect of our various approximations employed to compute the scene geometry on the reliability of 3D gaze estimates derived from LAEO data.

\vspace{-4mm}
\paragraph{Geometric 2D LAEO Loss}
\label{sec:geom_2d}

For two subjects $A$ and $B$ in LAEO, the projections of their predicted 3D gaze vectors onto the scene image plane, should lie along the line joining their 2D cyclopean eyes $P^{2D}_A$ and $P^{2D}_A$ (see Fig.~\ref{fig:summary_methods}). This intuition forms the basis of our geometric 2D LAEO loss $L^{2D}_{geom}$. To compute this loss, we estimate the gaze angles $\hat{g}_A$ for subject $A$ in LAEO by forward propagating their head crop image $I_A$ through $\mathcal{F(\bf{I}, \theta)}$. We then transform it to a 3D unit gaze vector $\hat{g}^{3D}_A$ originating from subject $A$'s 3D cyclopean eye $P^{3D}_A$ in the camera coordinate system. Next, we forward project $\hat{g}^{3D}_A$ onto the observed scene image as the 2D gaze vector $\hat{g}^{2D}_A$ (see Fig.~\ref{fig:summary_methods}). To compute $L^{2D}_{geom}$, we compute the angular cosine distance between two 2D unit vectors in the image plane: one along $\hat{g}^{2D}_A$ and another one along the line joining  $P^{2D}_A$ and $P^{2D}_B$. We repeat this process for subject $B$ and average both losses to obtain the final loss $L^{2D}_{geom}$.

Note, however, that $L^{2D}_{geom}$ on its own cannot fully resolve the depth ambiguity present in videos obtained from the Internet and hence is not sufficient to learn 3D gaze (see Table~\ref{tab:best_loss}), but when combined with the other LAEO losses it helps to improve overall gaze estimation accuracy (see Sec.~B.4 in supplementary). 
Thus, we additionally propose a geometric 3D LAEO loss which helps to resolve depth ambiguities and aids in learning 3D gaze more accurately. We describe it next.

\vspace{-3mm}
\paragraph{Geometric 3D LAEO Loss}
\label{sec:geom_3d}
The geometric 3D LAEO loss, $L^{3D}_{geom}$, explicitly provides 3D directional information to supervise gaze learning. We formulate it to enforce that the estimated 3D gaze vector originating from the cyclopean eye $P^{3D}_B$ of subject $B$ in LAEO, must intersect the viewed subject $A$'s 3D cyclopean eye $P^{3D}_A$ (see Fig.~\ref{fig:summary_methods}). To achieve this, we first estimate the 3D facial plane $\Pi_A$ of the viewed subject $A$, and place it at their 3D cyclopean eye location $P^{3D}_A$ perpendicular to their heading vector. We define the heading vector as the line joining the 3D midpoint of a subject's outer most 3D ear points, and 3D nose tip obtained from the fitted SMPL head model. Then the geometric 3D LAEO constraint for subject $B$ is given by $||\hat{P}^{3D}_A$ - $P^{3D}_A||$, where $P^{3D}_A$ is subject $A$'s 3D cyclopean eye position and $\hat{P}^{3D}_A$ is the intersection of subject $B$'s 3D gaze vector $\hat{g}^{3D}_B$ with subject $A$'s face plane $\Pi_A$ (see Fig.~\ref{fig:summary_methods}). Here $||.||$ denotes Euclidean distance. We repeat this process for subject $A$ and average the losses computed for both subjects to obtain the final loss $L^{3D}_{geom}$. Empirically we find that our formulation for $L^{3D}_{geom}$ performs better than an alternate cosine angle-based version (see Sec.~B.3 in supplementary).

\vspace{-3mm}
\paragraph{Pseudo Gaze LAEO Loss}
\label{sec:laeo_loss}
The LAEO activity also provides us with the self-supervised constraint that the ground truth 3D gaze vectors of two individuals $A$ and $B$ in LAEO, are oriented along the same 3D line, but in opposite directions to each other, \textit{i.e.}, $g^{3D}_A=-g^{3D}_B$. Hence, we leverage it in a self-training procedure and compute gaze pseudo ground truth labels for a pair of LAEO subjects continually while training. We observe that the LAEO activity often results in a clear frontal view of one subject while the other subject is turned away (see examples in Fig.~\ref{fig:teaser} and Fig.~\ref{fig:summary_methods}). Moreover, gaze estimation errors generally increase with extreme head poses where features such as the eyes are less visible (see Fig.~2 in the supplementary for a plot of gaze error versus gaze yaw). 
For example, in the extreme case of looking from behind a subject, facial features become completely occluded. 

We find that the uncertainty measure estimated by our network is well correlated with gaze error (with a Spearman's rank correlation coefficient of value of $0.46$). So to derive the gaze pseudo ground truth for a pair of faces in LAEO, we use the uncertainty measure to weigh more heavily the more reliable (less uncertain) of the two gaze estimates for a LAEO pair. Specifically, let $\{\hat{g}^{3D}_A, \hat{\sigma}_A\}$ and $\{\hat{g}^{3D}_B, \hat{\sigma}_B\}$ be the predicted 3D gaze vectors and their angular uncertainty values (in a common 3D coordinate system) for a pair of input face crops in LAEO, $\bf{I_A}$ and $\bf{I_B}$, respectively. We compute the pseudo 3D gaze ground truth label $g^{3D}_{pseudo}$ for faces $A$ and $B$ as a weighted combination of their estimated 3D gaze vectors as:
\vspace{-2mm}
\begin{equation}
    g^{3D}_{pseudo} = \mathrm{w_A} \hat{g}^{3D}_A\ + \mathrm{w_B} (-\hat{g}^{3D}_B),
    \vspace{-0.5mm}
\end{equation}

where we compute $\mathrm{w_A}$ and $\mathrm{w_B}$ from the angular uncertainty values $\hat{\sigma}_A$ and $\hat{\sigma}_B$ as $\mathrm{w_A}=\nicefrac{\hat{\sigma}_B}{\hat{\sigma}_A + \hat{\sigma}_B}$ and $\mathrm{w_B}=\nicefrac{\hat{\sigma}_A}{\hat{\sigma}_A + \hat{\sigma}_B}$ predicted by the gaze network. We further compute cosine distances between each LAEO subjects' predicted gaze vectors $\hat{g}^{3D}$ and their respective pseudo ground truth values $g^{3D}_{pseudo}$ and $-g^{3D}_{pseudo}$. We average the cosine distances computed for both subjects to obtain the final $L^{pseudo}_G$ loss. We find that this formulation of $L^{pseudo}_G$ is superior to other variants of it (see Sec.~B.2 in supplementary). 

\vspace{-3mm}
\paragraph{Aleatoric Gaze Loss}
\label{sec:aleatoric} We use an aleatoric loss function $L_G$ to supervise gaze estimation of individual face inputs, which regresses both the predicted gaze value and its uncertainty. This gaze uncertainty is helpful in deriving pseudo ground truths for pairs of faces in LAEO as described in the previous section. Aleatoric uncertainty models the distribution of the estimated gaze angles as a parametric Laplacian function and hence our gaze network $\mathcal{F(\bf{I}, \theta)}$ predicts their estimated mean $\{\hat{g}_{\theta}, \hat{g}_{\phi}\}$ and absolute deviation $\hat{\sigma}$ values. We supervise the network by minimizing the negative log-likelihood of observing the ground truth gaze value $\{g_{\theta}, g_{\phi}\}$ w.r.t.\ to this predicted Laplacian distribution as:
\vspace{-2mm}
\begin{align}
    L^\theta_G &= \mathrm{log}(\hat{\sigma}) + \frac{1}{\hat{\sigma}}|{\hat{g}_\theta - g_\theta}| \nonumber \\ 
    L^\phi_G &= \mathrm{log}(\hat{\sigma}) + \frac{1}{\hat{\sigma}}|{\hat{g}_\phi - g_\phi}|  \\
    L_G &= L^\phi + L^\theta \nonumber.
    \vspace{-4mm}
\end{align}

\noindent In practice, we predict the logarithm of the absolute deviation $\mathrm{log}(\hat{\sigma})$ from our network. This formulation has been shown to be numerically stable and avoids a potential division by zero~\cite{Kendall2017}. Note that previously, in~\cite{Kellnhofer2019}, the authors similarly employed a pinball loss to estimate the uncertainty of gaze predictions. We find that, in comparison to the pinball loss, the aleatoric loss improves the baseline accuracy of gaze estimation (see Sec.~B.1 in supplementary).

\vspace{-2mm}
\paragraph{Symmetry Loss} We also exploit the left-right symmetry inherent to the gaze estimation task to enforce another self-supervised gaze symmetry loss $L_{sym}$. Specifically, we estimate gaze angles for an input face image $\bf{I}$ as $\hat{g} = \{\hat{g}_{\theta}, \hat{g}_{\phi}\}$, reverse the sign of its predicted gaze yaw angle to produce the altered prediction $\hat{g}^* = \{\hat{g}_{\theta}, -\hat{g}_{\phi}\}$ and use this altered gaze estimate as the ground truth to supervise the predicted gaze, using the aleatoric loss, for a horizontally flipped (mirrored) version $\bf{I^*}$ of the input face image as:
\begin{equation}
\vspace{-0.5mm}
    L_{sym} = L_G(\mathcal{F(\bf{I^*}, \theta)},\hat{g}^*).
    \vspace{-0.5mm}
\end{equation}

We repeat this process for the horizontally flipped image and average the two resultant losses. Note that here the gaze angles are assumed to be in a normalized eye coordinate system as described in~\cite{Kellnhofer2019}, whose $z$ axis passes through each subject's 3D cyclopean eye position $P^{3D}$. This loss prevents network over-fitting while improving accuracy of gaze estimation (see Sec.~B.1 in supplementary).

\subsection{Training}
\label{sec:methods:training}
\vspace{-2mm}
We adopt two training paradigms: purely weakly-supervised training with LAEO data only or semi-supervised training where LAEO data augments data containing explicit 3D gaze labels. In both conditions, we initialize the ResNet-18 backbone of our model with weights pre-trained using ImageNet~\cite{imagenet_cvpr09}. We initialize the LSTM module and FC weights using a normal distribution. For semi-supervised training, we first train our model to convergence with images containing explicit 3D gaze labels only and then add weakly-supervised images with LAEO labels and continue training jointly to convergence.
We fix the parameters of the batch normalization layers during initialization to those found in the ImageNet pre-trained weights. We optimize the model using the following objective function:
\vspace{-1mm}
\begin{equation}
\begin{split}
    L &= L_{G} + \alpha L_{sym} + \beta L_{LAEO},\\
    L_{LAEO} &= (L^{3D}_{geom} + L^{2D}_{geom} + L^{pseudo}_G).
\end{split}
\vspace{-1mm}
\end{equation}

Here, $\alpha$ and $\beta$ are scalar weights, which slowly ramp up the contribution of the symmetry and LAEO losses, respectively. The ramp operation is formulated as ($\ceil {i/T}^1$) where $i$ is the smallest iterative step to update our model while $T$ is a threshold. We set $T_\alpha$ as 3000 and $T_\beta$ as 2400. In experiments, which do not involve any gaze supervision, $\beta$ is always fixed at 1 and $L_G$ is not included. We use a batch-size of 80 frames/sequences to train our static/temporal gaze estimation network. We use a fixed learning rate of $10^{-4}$ with the ADAM optimizer~\cite{kingma2014adam}.

\section{Experiments}
\vspace{-2mm}
Here we evaluate the real-world performance of our method in the fully weakly-supervised or semi-supervised settings for the task of physically unconstrained gaze estimation~\cite{Kellnhofer2019}. We perform extensive experiments within and across datasets. Besides gaze estimation, in Sec.~A of the supplementary, we also show the utility of adding LAEO labels to the task of in-the-wild visual target attention prediction~\cite{Chong2020} in a semi-supervised setting. 

\begin{figure*}
    \centering
    \includegraphics[width=0.9\linewidth]{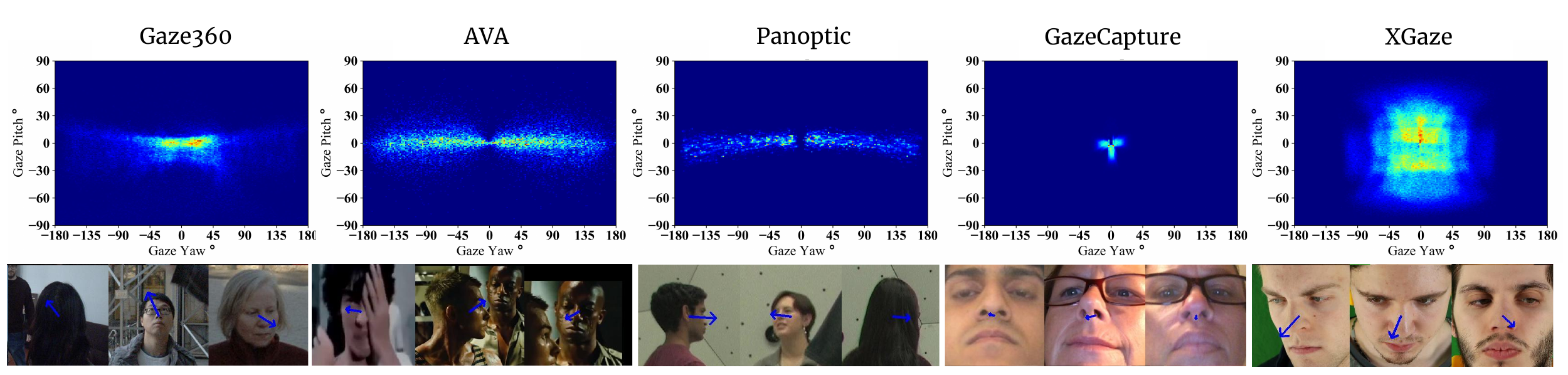}
    \caption{\textbf{Top} Gaze direction distribution of the Gaze360~\cite{Kellnhofer2019}, AVA-LAEO~\cite{Marin-Jimenez2019,Gu2018}, CMU Panoptic~\cite{joo2015panoptic}, GazeCapture~\cite{Krafka2016CVPR} and ETH-XGaze~\cite{Zhang2020ETHXGaze} datasets. Note that here the approximate gaze for CMU Panoptic and AVA-LAEO is computed by joining the LAEO pair of subjects' 3D cyclopean eye locations. \textbf{Bottom} Example face or head crops (if available) from the individual datasets.}
    \label{fig:gaze_dist}
    \vspace{-2mm}
\end{figure*}

\vspace{-2mm}
\paragraph{LAEO Datasets} We employ two LAEO datasets -- CMU Panoptic~\cite{Joo2019} and AVA~\cite{Gu2018,Marin-Jimenez2019}. CMU Panoptic~\cite{joo2015panoptic} is collected with a multiple-camera system installed in a large indoor dome, wherein subjects perform various activities. It does not contain LAEO annotations but contains the subjects' 3D body-joint locations and camera calibration information, which we directly use in our experiments. From video sequences containing the \textit{haggling} activity, we extract clips with the LAEO activity present via a semi-automatic procedure (described Sec.~D.1 of supplementary). This results in over 800k pairs of faces extracted from 485 unique subjects. For our experiments, we only utilize images from cameras that are parallel to the ground plane.

To acquire LAEO data from in-the-wild Internet videos, we leverage the large scale AVA human activity dataset~\cite{Gu2018} with LAEO annotations~\cite{Marin-Jimenez2019} provided by Marin-Jimenez~\etal (called ``AVA-LAEO"). It consists of annotated head bounding box pairs under LAEO in select frames across multiple video sequences, resulting in a wide variety of faces, backgrounds and lighting conditions. Unlike CMU Panoptic, AVA-LAEO does not provide access to accurate camera parameters or 3D human poses. We estimate the subjects' 3D poses using DensePose~\cite{Guler2018} and AlphaPose~\cite{fang2017rmpe} (described in Sec.~\ref{sec:scene_geometry} and Sec.~D.2 in supplementary). In all, this dataset contains 13,787 sequences of pairs of faces in LAEO.

\vspace{-5mm}
\paragraph{Gaze Datasets} We validate the efficacy of our weakly-supervised approach on the large-scale physically unconstrained in-the-wild Gaze360~\cite{Kellnhofer2019} dataset. It contains explicit 3D gaze labels and large variations in subject head poses and gaze angles, and lighting conditions and backgrounds. Its images are acquired in both indoor and outdoor environments using a Ladybug multi-camera system. It contain 127K training sequences from 365 subjects. For semi-supervised training, we additionally use two large-scale gaze datasets with known 3D gaze ground-truth --  GazeCapture~\cite{Krafka2016CVPR} and ETH-XGaze~\cite{Zhang2020ETHXGaze}. GazeCapture contains nearly 2M frontal face images of 1474 subjects acquired in unconstrained environmental conditions. ETH-XGaze, on the other hand, was acquired indoors with controlled lighting on a standard green background with a multi-view camera system. It contains 756K frames of 80 subjects.

The gaze distribution plots of all these datasets and their example face images are shown in Fig.~\ref{fig:gaze_dist}.
For GazeCapture and ETH-XGaze, we use the normalization procedure described in~\cite{Zhang2017CVPRW} to create normalized face crops. For all other datasets, we employ the procedure described in~\cite{Kellnhofer2019} to create normalized head crops. For all evaluations, we report the angular error (in degrees) between the estimated and ground truth unit gaze vectors, averaged across the corresponding test dataset.

\vspace{-2mm}
\subsection{Ablation Study }
\label{sec:ablation_study}
\vspace{-2mm}
To verify the contributions of our individual losses, we conduct a purely weakly-supervised cross-dataset ablation study. We train our method with the CMU Panoptic or AVA-LAEO datasets and evaluate performance on the Gaze360 dataset's test partition. Table~\ref{tab:best_loss} highlights the effect of the various weakly-supervised LAEO losses in this cross-dataset setting. All values reported are for the case when the symmetry loss was used by default. We train two configurations of our gaze estimation model -- (a) a temporal version, which accepts 7 frames as input, and (b) a static variant, which predicts gaze from a single input frame.

We observe that among the individual weakly-supervised losses, $L^{pseudo}_{G}$ and $L^{2D}_{geom}$ on their own or together, result in degenerate solutions. This is not surprising as it highlights the effects of depth ambiguity (see Sec.~\ref{sec:geom_2d}). Strong supervision can be provided by explicitly constraining the estimated gaze to intersect a 3D target, in our case, the viewed subject's head in the LAEO condition. This can be seen from the fact that $L^{3D}_{geom}$ significantly improves over its degenerate counterparts. We observe that the best performance is achieved by utilizing a combination of $L^{3D}_{geom}$, $L^{2D}_{geom}$ and the $L^{pseudo}_{G}$ losses, especially with the real-world AVA-LAEO dataset where the scene geometry is not known. We also find that removing the symmetry loss increases the overall gaze error of our best (temporal) model to $27.9^{\circ}$ from $25.9^{\circ}$ for CMU Panoptic and to $27.9^{\circ}$ from $26.3^{\circ}$ for AVA-LAEO (not listed in Table~\ref{tab:best_loss}). We provide additional ablation studies to explore the effect of the aleatoric and symmetric losses; other variants of the $L^{pseudo}_{G}$ and $L^{3D}_{geom}$ losses; and the utility of $L^{2D}_{geom}$ in Sec.~B of the supplementary.

\renewcommand{\arraystretch}{1.15}
\begin{table}[]
\centering
\small
\begin{tabular}{l|ll|lr}
\hline
 & \multicolumn{2}{c|}{Temporal} & \multicolumn{2}{c}{Static} \\ \hline
\begin{tabular}[c]{@{}l@{}}Loss functions\end{tabular} & \multicolumn{1}{l}{Pano.} & \multicolumn{1}{l|}{AVA} & \multicolumn{1}{l}{Pano.} & \multicolumn{1}{l}{AVA} \\ \hline
$L^{pseudo}_{G}$ & 55.4 & 52.9 & 61.9 & 48.0 \\
$L^{2D}_{geom}$ & 58.7 & 52.4 & 49.0 & 46.9 \\
$L^{3D}_{geom}$ & 28.0 & 30.1 & 31.4 & 30.4 \\ \hline
$L^{2D}_{geom}+L^{pseudo}_{G}$ & 55.0 & 54.1 & 54.0 & 51.3 \\
$L^{3D}_{geom}+L^{pseudo}_{G}$ & 26.1 & 27.3 & \textbf{29.0} & 30.6 \\
$L^{3D}_{geom}+L^{2D}_{geom}$ & 26.9 & 26.4 & 31.3 & 30.8 \\ \hline
\begin{tabular}[c]{@{}l@{}}$L^{pseudo}_{G}+L^{3D}_{geom}+L^{2D}_{geom}$ \end{tabular} & \multicolumn{1}{l}{\textbf{25.9}} & \multicolumn{1}{l|}{\textbf{26.3}} & \multicolumn{1}{l}{31.3} & \textbf{28.7} \\ \hline
\end{tabular}%
\vspace{1mm}
\caption{An ablation study to evaluate our individual weakly-supervised LAEO losses. The symmetry loss is always used. All numbers reported are using predictions from the temporal and static variants of our gaze estimation model, when evaluated on Gaze360's test set, measured in gaze angular error in degrees. Lower is better.
}
\label{tab:best_loss}
\vspace{-0.5cm}
\end{table}

\subsection{Semi-supervised Evaluation}
\vspace{-2mm}
Despite successfully learning to estimate gaze, the performance of our purely weakly-supervised model (trained on the AVA-LAEO dataset and tested on the Gaze360 dataset) lags behind the fully-supervised model on Gaze360's training set~\cite{Kellnhofer2019} as shown in Table~\ref{tab:cross_dataset} ($26.3^{\circ}$ vs $13.2^{\circ}$ for the temporal model). One reason for this discrepancy is the presence of noise in the gaze labels derived from LAEO data (as discussed in Sec.~D.3 of the supplementary) and the other is the
the existence of domain gap between the AVA-LAEO and Gaze360 datasets. The latter is evident from the gaze distribution plots shown in Fig.~\ref{fig:gaze_dist}. LAEO data tends to be biased towards viewing individuals from larger profile angles (see Fig.~\ref{fig:teaser} and Fig.~\ref{fig:summary_methods}) and contains less frontal face data. It also contains less diversity in the head's pitch (up/down rotation).

Hence, in this experiment, we explore a semi-supervised setting, where we evaluate if weakly-supervised LAEO data can successfully augment limited gaze-labeled data and improve its generalization for the task of physically unconstrained gaze estimation in-the-wild. We conduct both cross-dataset and within-dataset experiments. For the cross-dataset experiment, we train our model with several existing datasets other than Gaze360 and test on Gaze360's test partition. For the within-dataset experiment, we train on various subsets of Gaze360's training partition along with LAEO data and evaluate performance on Gaze360's test set. Unlike~\cite{Zhang2020ETHXGaze}, which evaluates performance on only frontal faces from Gaze360, we evaluate performance on both (a) frontal and (b) all faces from Gaze360's test set (including large profile faces).

\begin{table}[t]
\centering
\small
\begin{tabular}{l|cc}
\hline
\multicolumn{3}{c}{\bf Within dataset, Gaze + LAEO Labels} \\
 \hline
Training Data & Frontal face crops & All head crops \\ \hline
Gaze360~\cite{Kellnhofer2019} & 11.1 & 13.5 \\
Gaze360 & \textbf{10.1} & \textbf{13.2} \\
Gaze360 + AVA & 10.2 & \textbf{13.2} \\
\hline
\multicolumn{3}{c}{\bf Cross dataset, Gaze + LAEO Labels} \\
\hline
Training Data & Frontal face crops & All head crops \\ \hline
GazeCapture~\cite{Zhang2020ETHXGaze}  & 30.2 & -  \\
GazeCapture  & 29.2 & 58.2  \\
GazeCapture + AVA & 19.5 & 27.2 \\
\hline
ETH-XGaze~\cite{Zhang2020ETHXGaze} & 27.3 & -  \\
ETH-XGaze & 20.5 & 52.6  \\
ETH-XGaze + AVA & {\bf 16.9} & {\bf 25.0} \\
\hline
\hline
\multicolumn{3}{c}{\bf Cross dataset, LAEO Labels} \\
\hline
Training Data & Frontal face crops & All head crops \\
\hline
AVA & 29.0 & 26.3 \\
CMU Panoptic & 26.0 & 25.9  \\
CMU Panoptic + AVA & {\bf 22.5} & {\bf 24.4} \\
\hline
\end{tabular}%
\vspace{1mm}
\caption{Performance evaluation of our temporal model on Gaze360 dataset's test partition with various different training datasets ranging from those containing full gaze supervision (Gaze360, GazeCapture, ETH-XGaze), weak LAEO supervision only (the AVA-LAEO or CMU Panoptic datasets), or their combinations. All reported values are gaze angular errors in degrees (lower is better) on either (a) frontal face crops only or (b) all head crops from Gaze360's test set. Note that the addition of AVA-LAEO to GazeCapture or ETH-XGaze significantly improves their generalization performance on Gaze360.}
\label{tab:cross_dataset}
\vspace{-0.5cm}
\end{table}

\vspace{-0.5cm}
\paragraph{Cross-dataset} In Table~\ref{tab:cross_dataset}, we compare the generalization performance of the GazeCapture and the ETH-XGaze datasets on Gaze360, with and without weak supervision from AVA-LAEO. Both these supervised gaze datasets, although large, are limited in some respect for the task of physically unconstrained gaze estimation in-the-wild. The GazeCapture dataset contains images acquired indoors and outdoors, but of mostly frontal faces with a narrow distribution of gaze angles (Fig.~\ref{fig:gaze_dist}). The ETH-XGaze dataset, on the other hand, has a broad distribution of gaze angles from 80 subjects (Fig.~\ref{fig:gaze_dist}), but is captured indoors only.

Table~\ref{tab:cross_dataset} highlights that on including weak gaze supervision from AVA-LAEO, the generalization performances of both  GazeCapture and ETH-XGaze on Gaze360, for frontal and all faces is improved. For frontal faces, the addition of AVA-LAEO results in improvements of $7.4^\circ$ for GazeCapture and $3.6^\circ$ for ETH-XGaze. On all head crops, however, this improvement is even more pronounced -- $31.0^\circ$ for GazeCapture and $27.6^\circ$ for ETH-XGaze. Fig.~\ref{fig:gaze_dist} shows that the AVA-LAEO dataset complements both the GazeCapture and ETH-XGaze datasets by expanding their underlying distributions via weak gaze labels (see more details in Sec.~C of supplementary). In Table~\ref{tab:cross_dataset}, we also show the cross-dataset performance of jointly training with CMU Panoptic and AVA-LAEO with their weak gaze labels only. We find that the in-the-wild AVA-LAEO data also slightly improves the generalization performance of the indoor-only CMU Panoptic data on Gaze360. Finally, Table~\ref{tab:cross_dataset} shows that our model also outperforms the previously reported state-of-the-art performances~\cite{Kellnhofer2019, Zhang2020ETHXGaze} on all benchmarks. 

\begin{figure}
    \centering
    \includegraphics[width=1.0\linewidth]{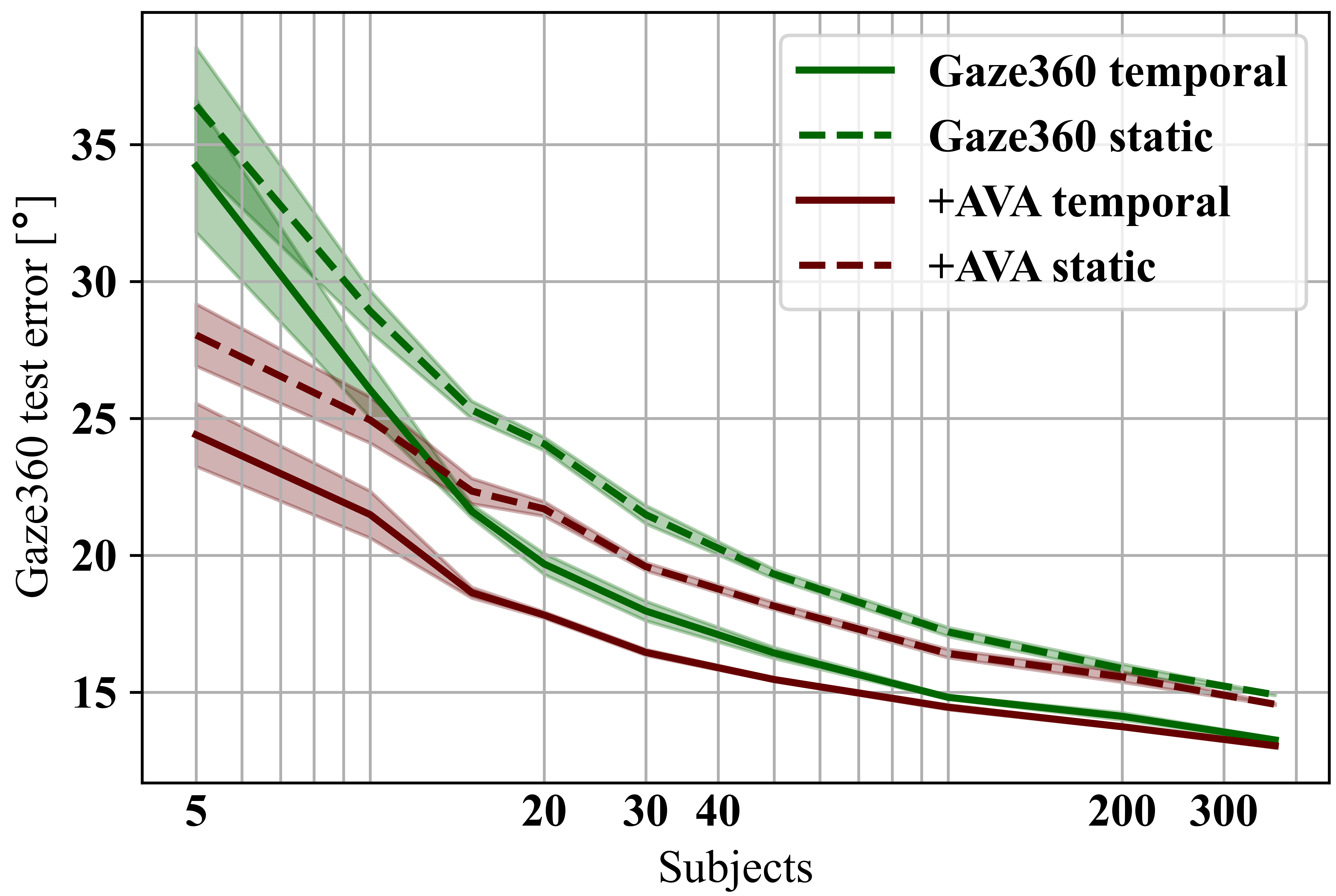}
    \caption{Gaze error (in degrees) on the Gaze360 test set on augmenting a reduced Gaze360 training set (with less subjects) with AVA-LAEO. We vary the number of Gaze360 training subjects along the horizontal axis. The shaded area corresponds to the standard error of the average metric evaluated over 5 repetitions of each experiment performed by picking a different random subset of subjects each time.
    }
    \label{fig:tinygaze360_results}
    \vspace{-0.5cm}
\end{figure}

\vspace{-0.35cm}
\paragraph{Within-dataset}
Training data acquired from a larger number of subjects improves generalization of gaze estimators as shown in~\cite{Krafka2016CVPR}. 
However, recruiting more subjects requires additional cost and time. 
In Fig.~\ref{fig:tinygaze360_results}, we evaluate the performance of training with progressively larger numbers of subjects from Gaze360's training set, without (labeled as ``Gaze360" in Fig.~\ref{fig:tinygaze360_results}) and with (labeled as ``+AVA" in Fig.~\ref{fig:tinygaze360_results}) AVA-LAEO. 
We use all available videos of a particular subject during training. We assess both our temporal and static models.
For this within-domain semi-supervised setting, we find that training on a small number of subjects from Gaze360 along with weak supervision from AVA-LAEO, offers the same performance as using a larger number of subjects from Gaze360. 

\vspace{-0.1cm}
\section{Conclusion}
\vspace{-0.1cm}
In this work, we present the first exploration of a weakly-supervised 3D gaze learning paradigm from images/videos of people \textit{looking at each other} (LAEO). This approach is trivially scalable due to the ease of acquiring LAEO annotations from Internet videos. To facilitate the learning of 3D gaze, we propose three training objectives, which exploit the underlying geometry native to the LAEO activity. Through many experiments we demonstrate that our approach is successful in augmenting gaze datasets limited in gaze distributions, subjects, or environmental conditions with unconstrained images of people under LAEO, resulting in improved physically unconstrained gaze estimation in the wild.

{\small
\bibliographystyle{ieee_fullname}
\bibliography{egbib}

\begin{thebibliography}{10}\itemsep=-1pt

\bibitem{chen2020offset}
Zhaokang Chen and Bertram Shi.
\newblock Offset calibration for appearance-based gaze estimation via gaze
  decomposition.
\newblock In {\em WACV}, pages 270--279, 2020.

\bibitem{cheng2018appearance}
Yihua Cheng, Feng Lu, and Xucong Zhang.
\newblock Appearance-based gaze estimation via evaluation-guided asymmetric
  regression.
\newblock In {\em ECCV}, pages 100--115, 2018.

\bibitem{chong2018connecting}
Eunji Chong, Nataniel Ruiz, Yongxin Wang, Yun Zhang, Agata Rozga, and James~M
  Rehg.
\newblock Connecting gaze, scene, and attention: Generalized attention
  estimation via joint modeling of gaze and scene saliency.
\newblock In {\em ECCV}, pages 383--398, 2018.

\bibitem{Chong2020}
Eunji Chong, Yongxin Wang, Nataniel Ruiz, and James~M. Rehg.
\newblock {Detecting Attended Visual Targets in Video}.
\newblock In {\em CVPR}, 2020.

\bibitem{imagenet_cvpr09}
J. Deng, W. Dong, R. Socher, L.-J. Li, K. Li, and L. Fei-Fei.
\newblock {ImageNet: A Large-Scale Hierarchical Image Database}.
\newblock In {\em CVPR}, 2009.

\bibitem{deng2019accurate}
Yu Deng, Jiaolong Yang, Sicheng Xu, Dong Chen, Yunde Jia, and Xin Tong.
\newblock Accurate 3d face reconstruction with weakly-supervised learning: From
  single image to image set.
\newblock In {\em CVPR Workshops}, 2019.

\bibitem{fang2017rmpe}
Hao-Shu Fang, Shuqin Xie, Yu-Wing Tai, and Cewu Lu.
\newblock {RMPE}: Regional multi-person pose estimation.
\newblock In {\em ICCV}, 2017.

\bibitem{Fischer2018}
Tobias Fischer, Hyung Jin~Chang, and Yiannis Demiris.
\newblock Rt-gene: Real-time eye gaze estimation in natural environments.
\newblock In {\em ECCV}, pages 334--352, 2018.

\bibitem{FunesMora2014ETRA}
Kenneth~Alberto Funes~Mora, Florent Monay, and Jean-Marc Odobez.
\newblock Eyediap: A database for the development and evaluation of gaze
  estimation algorithms from rgb and rgb-d cameras.
\newblock In {\em ACM ETRA}. ACM, Mar. 2014.

\bibitem{Gu2018}
Chunhui Gu, Chen Sun, David~A. Ross, Carl Vondrick, Caroline Pantofaru, Yeqing
  Li, Sudheendra Vijayanarasimhan, George Toderici, Susanna Ricco, Rahul
  Sukthankar, Cordelia Schmid, and Jitendra Malik.
\newblock {AVA: A Video Dataset of Spatio-Temporally Localized Atomic Visual
  Actions}.
\newblock In {\em CVPR}, 2018.

\bibitem{Guler2018}
Riza~Alp G{\"{u}}ler, Natalia Neverova, and Iasonas Kokkinos.
\newblock {DensePose: Dense Human Pose Estimation in the Wild}.
\newblock In {\em CVPR}, 2018.

\bibitem{He_2019_ICCV}
Junfeng He, Khoi Pham, Nachiappan Valliappan, Pingmei Xu, Chase Roberts, Dmitry
  Lagun, and Vidhya Navalpakkam.
\newblock On-device few-shot personalization for real-time gaze estimation.
\newblock In {\em ICCV Workshops}, Oct 2019.

\bibitem{Huang2017MVA}
Qiong Huang, Ashok Veeraraghavan, and Ashutosh Sabharwal.
\newblock Tabletgaze: Dataset and analysis for unconstrained appearance-based
  gaze estimation in mobile tablets.
\newblock {\em Mach. Vision Appl.}, 28(5-6):445--461, Aug. 2017.

\bibitem{iqbal2020weakly}
Umar Iqbal, Pavlo Molchanov, and Jan Kautz.
\newblock Weakly-supervised 3d human pose learning via multi-view images in the
  wild.
\newblock In {\em CVPR}, pages 5243--5252, 2020.

\bibitem{joo2015panoptic}
Hanbyul Joo, Hao Liu, Lei Tan, Lin Gui, Bart Nabbe, Iain Matthews, Takeo
  Kanade, Shohei Nobuhara, and Yaser Sheikh.
\newblock Panoptic studio: A massively multiview system for social motion
  capture.
\newblock In {\em ICCV}, pages 3334--3342, 2015.

\bibitem{Joo2019}
Hanbyul Joo, Tomas Simon, Xulong Li, Hao Liu, Lei Tan, Lin Gui, Sean Banerjee,
  Timothy Godisart, Bart Nabbe, Iain Matthews, Takeo Kanade, Shohei Nobuhara,
  and Yaser Sheikh.
\newblock {Panoptic Studio: A Massively Multiview System for Social Interaction
  Capture}.
\newblock {\em TPAMI}, 2019.

\bibitem{Kellnhofer2019}
Petr Kellnhofer, Adria Recasens, Simon Stent, Wojciech Matusik, and Antonio
  Torralba.
\newblock {Gaze360: Physically unconstrained gaze estimation in the wild}.
\newblock {\em ICCV}, pages 6911--6920, 2019.

\bibitem{Kendall2017}
Alex Kendall and Yarin Gal.
\newblock {What uncertainties do we need in Bayesian deep learning for computer
  vision?}
\newblock In {\em NeurIPS}, pages 5575--5585, 2017.

\bibitem{kingma2014adam}
Diederik~P Kingma and Jimmy Ba.
\newblock Adam: A method for stochastic optimization.
\newblock {\em arXiv preprint arXiv:1412.6980}, 2014.

\bibitem{Krafka2016CVPR}
Kyle Krafka, Aditya Khosla, Petr Kellnhofer, Harini Kannan, Suchendra
  Bhandarkar, Wojciech Matusik, and Antonio Torralba.
\newblock {Eye Tracking for Everyone}.
\newblock In {\em CVPR}, June 2016.

\bibitem{lepetit2009epnp}
Vincent Lepetit, Francesc Moreno-Noguer, and Pascal Fua.
\newblock Epnp: An accurate o (n) solution to the pnp problem.
\newblock {\em IJCV}, 81(2):155, 2009.

\bibitem{linden2019learning}
Erik Lind{\'e}n, Jonas Sjostrand, and Alexandre Proutiere.
\newblock Learning to personalize in appearance-based gaze tracking.
\newblock In {\em ICCV Workshops}, pages 0--0, 2019.

\bibitem{liu2019differential}
Gang Liu, Yu Yu, Kenneth Alberto~Funes Mora, and Jean-Marc Odobez.
\newblock A differential approach for gaze estimation.
\newblock {\em TPAMI}, 2019.

\bibitem{SMPL:2015}
Matthew Loper, Naureen Mahmood, Javier Romero, Gerard Pons-Moll, and Michael~J.
  Black.
\newblock {SMPL}: A skinned multi-person linear model.
\newblock {\em SIGGRAPH Asia}, 34(6):248:1--248:16, Oct. 2015.

\bibitem{Marin-Jimenez2019}
Manuel~J. Marin-Jimenez, Vicky Kalogeiton, Pablo Medina-Suarez, and Andrew
  Zisserman.
\newblock {Laeo-net: Revisiting people looking at each other in videos}.
\newblock {\em CVPR}, 2019-June(i):3472--3480, 2019.

\bibitem{Marin-Jimenez2014}
M.~J. Marin-Jimenez, A. Zisserman, M. Eichner, and V. Ferrari.
\newblock {Detecting people looking at each other in videos}.
\newblock {\em IJCV}, 2014.

\bibitem{mazur1980physiological}
Allan Mazur, Eugene Rosa, Mark Faupel, Joshua Heller, Russell Leen, and Blake
  Thurman.
\newblock Physiological aspects of communication via mutual gaze.
\newblock {\em American Journal of Sociology}, 86(1):50--74, 1980.

\bibitem{Park2020ECCV}
Seonwook Park, Emre Aksan, Xucong Zhang, and Otmar Hilliges.
\newblock Towards end-to-end video-based eye-tracking.
\newblock In {\em ECCV}, 2020.

\bibitem{Park2019ICCV}
Seonwook Park, Shalini~De Mello, Pavlo Molchanov, Umar Iqbal, Otmar Hilliges,
  and Jan Kautz.
\newblock Few-shot adaptive gaze estimation.
\newblock In {\em ICCV}, 2019.

\bibitem{park2018deep}
Seonwook Park, Adrian Spurr, and Otmar Hilliges.
\newblock Deep pictorial gaze estimation.
\newblock In {\em ECCV}, pages 721--738, 2018.

\bibitem{park2018learning}
Seonwook Park, Xucong Zhang, Andreas Bulling, and Otmar Hilliges.
\newblock Learning to find eye region landmarks for remote gaze estimation in
  unconstrained settings.
\newblock In {\em ACM ETRA}, pages 1--10, 2018.

\bibitem{ranjan2018light}
Rajeev Ranjan, Shalini De~Mello, and Jan Kautz.
\newblock Light-weight head pose invariant gaze tracking.
\newblock In {\em CVPR Workshops}, pages 2156--2164, 2018.

\bibitem{recasens2015they}
Adria Recasens, Aditya Khosla, Carl Vondrick, and Antonio Torralba.
\newblock Where are they looking?
\newblock In {\em NeurIPS}, pages 199--207, 2015.

\bibitem{recasens2017following}
Adria Recasens, Carl Vondrick, Aditya Khosla, and Antonio Torralba.
\newblock Following gaze in video.
\newblock In {\em ICCV}, pages 1435--1443, 2017.

\bibitem{rhodin2018learning}
Helge Rhodin, J{\"o}rg Sp{\"o}rri, Isinsu Katircioglu, Victor Constantin,
  Fr{\'e}d{\'e}ric Meyer, Erich M{\"u}ller, Mathieu Salzmann, and Pascal Fua.
\newblock Learning monocular 3d human pose estimation from multi-view images.
\newblock In {\em CVPR}, pages 8437--8446, 2018.

\bibitem{Smith2013UIST}
B.A. Smith, Q. Yin, S.K. Feiner, and S.K. Nayar.
\newblock {G}aze {L}ocking: {P}assive {E}ye {C}ontact {D}etection for
  {H}uman-{O}bject {I}nteraction.
\newblock In {\em ACM UIST}, pages 271--280, Oct 2013.

\bibitem{spurr2020weakly}
Adrian Spurr, Umar Iqbal, Pavlo Molchanov, Otmar Hilliges, and Jan Kautz.
\newblock Weakly supervised 3d hand pose estimation via biomechanical
  constraints.
\newblock In {\em ECCV}, 2020.

\bibitem{striano2006social}
Tricia Striano and Vincent~M Reid.
\newblock Social cognition in the first year.
\newblock {\em Trends in cognitive sciences}, 10(10):471--476, 2006.

\bibitem{wang2018hierarchical}
Kang Wang, Rui Zhao, and Qiang Ji.
\newblock A hierarchical generative model for eye image synthesis and eye gaze
  estimation.
\newblock In {\em CVPR}, pages 440--448, 2018.

\bibitem{wang2019generalizing}
Kang Wang, Rui Zhao, Hui Su, and Qiang Ji.
\newblock Generalizing eye tracking with bayesian adversarial learning.
\newblock In {\em CVPR}, pages 11907--11916, 2019.

\bibitem{xiong2019mixed}
Yunyang Xiong, Hyunwoo~J Kim, and Vikas Singh.
\newblock Mixed effects neural networks (menets) with applications to gaze
  estimation.
\newblock In {\em CVPR}, pages 7743--7752, 2019.

\bibitem{yu2020unsupervised}
Yu Yu and Jean-Marc Odobez.
\newblock Unsupervised representation learning for gaze estimation.
\newblock In {\em CVPR}, pages 7314--7324, 2020.

\bibitem{Yu2020HUMBI}
Zhixuan Yu, Jae~Shin Yoon, In~Kyu Lee, Prashanth Venkatesh, Jaesik Park, Jihun
  Yu, and Hyun~Soo Park.
\newblock Humbi: A large multiview dataset of human body expressions.
\newblock In {\em CVPR}, pages 2990--3000, 2020.

\bibitem{Zhang2020ETHXGaze}
Xucong Zhang, Seonwook Park, Thabo Beeler, Derek Bradley, Siyu Tang, and Otmar
  Hilliges.
\newblock Eth-xgaze: A large scale dataset for gaze estimation under extreme
  head pose and gaze variation.
\newblock In {\em ECCV}, 2020.

\bibitem{zhang20_bmvc}
Xucong Zhang, Yusuke Sugano, Andreas Bulling, and Otmar Hilliges.
\newblock Learning-based region selection for end-to-end gaze estimation.
\newblock In {\em BMVC}, pages 1--13, 2020.

\bibitem{Zhang2015CVPR}
Xucong Zhang, Yusuke Sugano, Mario Fritz, and Andreas Bulling.
\newblock Appearance-based gaze estimation in the wild.
\newblock In {\em CVPR}, pages 4511--4520, June 2015.

\bibitem{Zhang2017CVPRW}
Xucong Zhang, Yusuke Sugano, Mario Fritz, and Andreas Bulling.
\newblock It's written all over your face: Full-face appearance-based gaze
  estimation.
\newblock In {\em CVPR Workshops}, pages 2299--2308, July 2017.

\bibitem{zhu2017monocular}
Wangjiang Zhu and Haoping Deng.
\newblock Monocular free-head 3d gaze tracking with deep learning and geometry
  constraints.
\newblock In {\em ICCV}, pages 3143--3152, 2017.

\end{thebibliography}
}

\clearpage

\setcounter{section}{0}
\renewcommand{\thesection}{\Alph{section}}
\section*{\Large Appendix}

In this supplementary document, we show additional experimental results and provide more implementation details. Specifically, we demonstrate the advantage of using weak labels from LAEO data on an additional in-the-wild physically unconstrained gaze-related task besides gaze estimation. For this we incorporate our gaze estimation pipeline from AVA-LAEO into the current state-of-the-art visual target estimation network~\cite{Chong2020} (termed ``VATnet" here) and evaluate its performance. Next, for the task of physically unconstrained gaze estimation, we provide additional ablation experiments (besides those in Sec.~4.1 of the main paper), including for the aleatoric and symmetry losses; for various formulations of the pseudo gaze and geometric 3D LAEO losses; and for the utility of the geometric 2D LAEO loss.
We show more performance details of the various training datasets used in the cross-dataset experiments (in Sec.~4.2 of the main paper) for different gaze yaw angles. Finally, we provide more details of pre-processing the CMU Panoptic and AVA-LAEO datasets, and analyze the reliability of the 3D gaze labels extracted from real-world LAEO data.

\section{Weakly-Supervised Visual Target Estimation}
\begin{figure}[h]
    \centering
    \includegraphics[width=\linewidth]{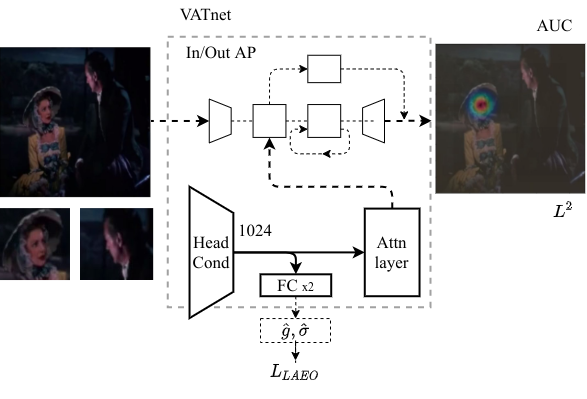}
    \caption{A simple modification of the VATnet architecture~\cite{Chong2020}. Two fully connected layers serve as an auxiliary task to predict 3D gaze from the head conditioning branch of the original VATnet architecture. The LAEO losses (see Section 3.3 in the main paper) on the predicted gaze vectors for the AVA-LAEO dataset are then used to fine-tune the final layer of the head conditioning branch. Facial features extracted from the fine-tuned head conditioning branch then proceed to VATnet for the visual attention target prediction task. Please refer to Chong~\etal~\cite{Chong2020} for a full description of their network architecture.}
    \label{fig:ava_eunji_arch}
\end{figure}

Chong \etal~\cite{Chong2020} proposed a novel spatio-temporal architecture (VATnet), which predicts fixation targets of subjects within a given video frame. In this experiment, we explore if LAEO-based weakly-supervised 3D gaze helps to estimate more accurate visual targets as well. We use LAEO 3D gaze estimation as an auxiliary task while training networks for visual target estimation in a semi-supervised setting. This provides additional weak gaze annotations from the noisy, in-the-wild AVA-LAEO dataset. 

\paragraph{Method}
VATnet comprises of four modules, a head conditioning branch, which generates gaze-related features from an input head image; a main scene branch, which generates scene-related feature maps based on the saliency of an input scene image; a recurrent attention prediction module, which fuses gaze- and scene-related features across contiguous video frames; and lastly, a heatmap conditioning branch, which generates a visual target prediction heatmap (see Fig.~\ref{fig:ava_eunji_arch}). 
VATnet's head conditioning branch is a ResNet-50 module initialized with weights from a gaze estimation network trained on the EYEDIAP dataset~\cite{FunesMora2014ETRA}. Utilizing this gaze estimator, Chong \etal~\cite{Chong2020} demonstrate state-of-the-art results on a new dataset called \textit{VisualAttentionTarget}, which comprises of annotated gaze target locations on the image plane. In our experiments we jointly train this VATNet architecture with both the training set of the original fully-supervised VAT dataset and with the AVA-LAEO dataset. To do so, we modify the VATNet architecture and add two fully connected layers to the output of the head conditioning branch, and train it to additionally predict weak 3D gaze vectors derived from the AVA-LAEO dataset (see Fig.~\ref{fig:ava_eunji_arch}). We train with samples from AVA-LAEO using the LAEO loss $L_{SYM} + L^{2D}_{geom} + L^{3D}_{geom} + L^{pseudo}_G$ only.

\paragraph{Data Preparation}
VATnet requires three input modalities. First, it requires a full scene image with known head bounding box locations for each annotated subject. Next, it requires a 2D pixel gaze target location on the image plane for the said subject and finally, an \textit{in-out} label, which indicates if the target is within or out of a frame. For this task, to use the LAEO data we input the same 7-frame sequence centered around a LAEO annotation. We treat the 2D cyclopean eye $P^{2D}$ (see the sub-section titled ``Scene Geometry Estimation'' within Sec.~3.3 of the main paper) of subject $B$ as the target for subject $A$ and vice versa for subject $B$. The nature of the AVA-LAEO data ensures that all target locations are within an image frame and we assume this to be the default \textit{in-out} ground truth state. We do not pre-process or augment the AVA-LAEO data and directly re-train Chong~\etal's original implementation of VATnet with the two datasets with minimal modifications.

\begin{table}[]
\centering
\resizebox{0.8\linewidth}{!}{%
\begin{tabular}{@{}c|ccc@{}}
\toprule
             & AUC ($\uparrow$)              & $L2$ Dist ($\downarrow$)     & \begin{tabular}[c]{@{}c@{}}\textit{out-of-frame}\\ AP ($\uparrow$)\end{tabular}  \\ \midrule
VAT          & 0.846             & 0.141           & \textbf{0.861}                                                              \\
VAT + AVA-LAEO & \textbf{0.865}             & \textbf{0.136}           & 0.855                                                              \\ \midrule
Human        & 0.921             & 0.051           & 0.925                                                              \\ \bottomrule
\end{tabular}
}
\vspace{0.3cm}
\caption{Improvements to the VATnet baseline~\cite{Chong2020} by adding weak supervision from the AVA-LAEO dataset using the best configuration of LAEO loss functions described in Table 1 of the main paper.}

\label{tab:eunji_ava}
\end{table}

\paragraph{Results} Following Chong~\etal~\cite{Chong2020}, we evaluate the area under the curve (AUC) for correct target location prediction (within a pre-specified distance threshold on the image plane), the $L2$ distance between the predicted and ground truth target locations in the scene and the out-of-frame prediction's average precision (AP). We report the scores on the VAT test dataset, averaged across training epochs 2-30, both for the author's original method~\cite{Chong2020} and our proposed modification. Table~\ref{tab:eunji_ava} shows the benefits of jointly training with the AVA-LAEO and VAT datasets. We notice an improvement in the AUC and $L2$ distance metrics for visual target prediction. These encouraging results suggest that weak supervision from noisily-labeled in-the-wild LAEO data can potentially also aid other gaze-related tasks, \textit{e.g.}, visual attention target prediction besides 3D gaze estimation. We also note a reduction in the \textit{out-of-frame} AP, which is not surprising as all target locations for a given subject in the AVA-LAEO dataset lie within image bounds and hence it provides labels for only one (\textit{i.e.}, the in-frame) class.

\section{Additional Ablation Studies}
For the task of physically unconstrained gaze estimation, we provide additional ablation experiments besides those in Sec.~4.1 of the main paper.

\subsection{Aleatoric and Symmetry Losses}
In the normalized eye co-ordinate system~\cite{Kellnhofer2019}, where the $z$ axis passes through the 3D cyclopean eye center of each face, constraining gaze yaw prediction to be equal and opposite for a face and its symmetrically flipped version, is an intuitive constraint, which can be employed during training. Our experiments show that using this symmetry constraint and the aleatoric gaze loss improve the baseline performance of~\cite{Kellnhofer2019} on both variants of the author's original fully-supervised ResNet-18-based gaze estimator (temporal and static), which use the pinball gaze loss. Table~\ref{tab:aleatoric_vs_pinball} shows a detailed comparison of the effects of adding the symmetry constraint to the pinball (from~\cite{Kellnhofer2019}) and aleatoric (ours) loss functions for a within-dataset fully-supervised experiment on Gaze360. Here we train our gaze network with Gaze360's entire training set (with its gaze labels) and evaluate it on Gaze360's test set. Note that the symmetry constraint improves the performance of both the pinball and aleatoric losses.

\begin{table}[]
\centering
\resizebox{\linewidth}{!}{%
\begin{tabular}{@{}c|cc|cc@{}}
\toprule
 & \multicolumn{2}{c|}{Temporal} & \multicolumn{2}{c}{Static} \\ \midrule
 & \begin{tabular}[c]{@{}c@{}}Frontal face\\ crops $\degree$\end{tabular} & \begin{tabular}[c]{@{}c@{}}All head\\ crops $\degree$\end{tabular} & \begin{tabular}[c]{@{}c@{}}Frontal face\\ crops $\degree$\end{tabular} & \begin{tabular}[c]{@{}c@{}}All head\\ crops $\degree$\end{tabular} \\ \midrule
Pinball & 10.38 & 13.77 & 11.4 & 15.62 \\
Aleatoric & 9.8 & 13.65 & 11.14 & 15.24 \\
Pinball$_{+L_{sym}}$ & 10.05 & 13.37 & 11.04 & 15.35 \\
Aleatoric$_{+L_{sym}}$ & \textbf{9.79} & \textbf{12.94} & \textbf{10.94} & \textbf{15.07} \\ \bottomrule
\end{tabular}%
}
\vspace{0.3cm}
\caption{Summary of performance gain by employing an aleatoric gaze loss (described in Sec.~3.3, ``Aleatoric Gaze Loss" of the main paper) and the effects of incorporating a symmetry constraint (described in Sec.~3.3, ``Symmetry Loss" of the main paper). All values reported are angular gaze errors in degrees (lower is better) for the fully-supervised within-dataset experiment on Gaze360.}
\label{tab:aleatoric_vs_pinball}
\end{table}

We also observe that for this within-dataset experiment, the aleatoric loss consistently outperforms the pinball loss and that the combination of the aleatoric and symmetry losses results in the best overall performance (Table~\ref{tab:aleatoric_vs_pinball}).
In addition to this, we observe that the aleatoric loss also outperforms the pinball loss in the cross-domain purely weakly-supervised experimental setting. By replacing the aleatoric loss with the pinball loss (from~\cite{Kellnhofer2019}), our best temporal network (trained with all the LAEO losses and corresponding to the last row of Table 1 in the main paper), generalizes less effectively to Gaze360. For AVA-LAEO its gaze error of 26.3$^\circ$ increases to 28.7$^\circ$ and for CMU Panoptic it increases from 25.9$^\circ$ to 26.1$^\circ$.

\begin{figure}
    \centering
    \includegraphics[width=\linewidth]{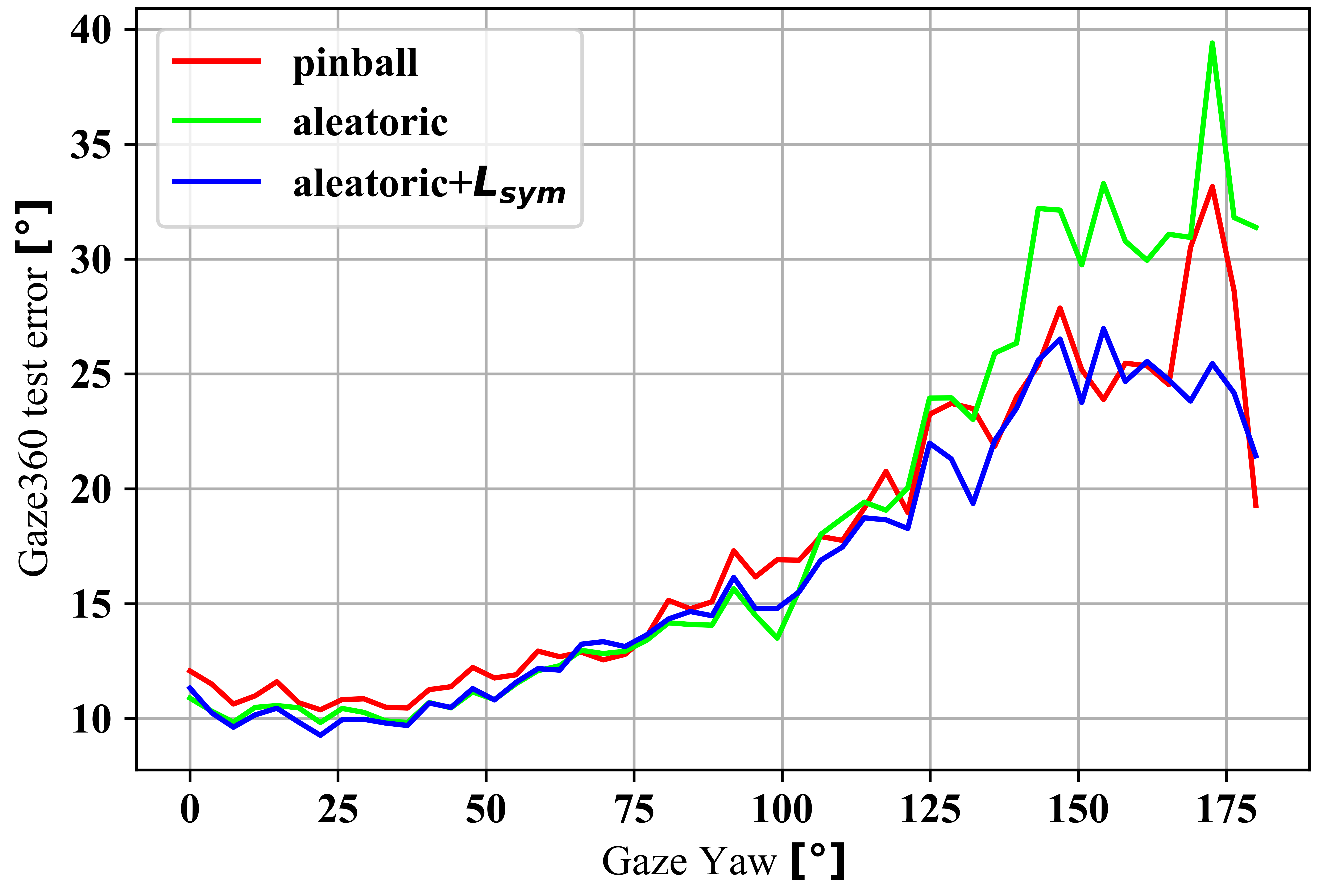}
    \caption{Gaze360 test error (in degrees) as a function of gaze yaw for the fully-supervised within-dataset experimental on Gaze360. Note that gaze error increases as faces turn away from the camera.}
    \label{fig:err_vs_yaw}
\end{figure}


\subsection{Variants of $L^{pseudo}_G$}
The LAEO activity provides us with the constraint that the predicted 3D gaze from subjects $A$ and $B$ in LAEO must be equal and opposite in a shared camera coordinate system. There are multiple ways in which we can implement this constraint. As an ablation, we explore two additional formulations for this LAEO constraint besides the one described in Sec.~3.3 titled ``Pseudo Gaze LAEO Loss" of the main paper: a) naive LAEO enforcement and b) using the most confident gaze prediction for a pair of faces in LAEO as the pseudo ground truth gaze direction. In either experiment, we replace the $L^{pseudo}_G$ loss in our best (temporal) purely weakly-supervised cross-dataset configuration that is trained with all the LAEO losses $L_{sym} + L^{pseudo}_G + L^{2D}_{geom} + L^{3D}_{geom}$ (corresponding to the last row in Table 1 of the main paper) with one of these losses.
\paragraph{Naive LAEO Enforcement} Here we naively enforce the predicted vectors $\hat{g}^{3D}_A$ and $\hat{g}^{3D}_B$ to be equal and opposite by minimizing the resultant angular cosine distance between $\hat{g}^{3D}_A$ and $-\hat{g}^{3D}_B$. In this constraint, predictions for both faces could be modified by the network. In order to achieve this, our gaze estimation network could either improve its prediction for the difficult face in a LAEO pair (see Fig.~\ref{fig:err_vs_yaw}, which show that gaze prediction error increased with extreme gaze angles), or it could deteriorate its prediction for the clearer frontal face to satisfy this naive LAEO objective. Our experiments show a reduction in cross-dataset performance on the entire Gaze360 test set (CMU Panoptic: $25.9^\circ$ $\rightarrow$ $28.2^\circ$ and AVA-LAEO: $ 26.3^\circ \rightarrow 26.9^\circ$) with this naive variant of the LAEO loss versus the one described in Sec.~3.3 of the main paper.
\paragraph{Confident Gaze Prediction} In this experiment, we regard the more confident of the two predicted gaze vectors for a LAEO pair as the pseudo ground truth $g^{3D}_{pseudo}$ gaze label as opposed to their weighted average used in Sec.~3.3 of the main paper. That is, $g^{3D}_{pseudo} = \hat{g}^{3D}_A$ if $W_A \geq W_B$ (from Eq. 1 in the main paper) and vice versa for subject $B$. Our experiments show a reduction in cross-dataset performance with this variant of the LAEO pseudo ground truth label as well versus the one used in Sec.~3.3 of the main paper (CMU Panoptic: $25.9^\circ$ $\rightarrow$ $27.24^\circ$ and AVA-LAEO: $ 26.3^\circ \rightarrow 27.8^\circ$).

\subsection{Variant of $L^{3D}_{geom}$}
We also compare the performance of our $L^{3D}_{geom}$ loss formulation used in Sec.~3.3 of the main paper to a conventional 3D angular cosine loss, whose ground truth is assumed to be along the line joining LAEO subjects' estimated 3D eyes. Empirically, we observe that replacing $L^{3D}_{geom}$ with a cosine loss in our best (temporal) purely weakly-supervised configuration (last row of Table~1 in the main paper), results in consistently worse performance on Gaze360 (CMU Panoptic: $25.9^\circ$ $\rightarrow$ $30.0^\circ$ and AVA-LAEO: $ 26.3^\circ \rightarrow 29.63^\circ$).

\subsection{Utility of $L^{2D}_{geom}$}
\begin{figure}
    \centering
    \includegraphics[width=\linewidth]{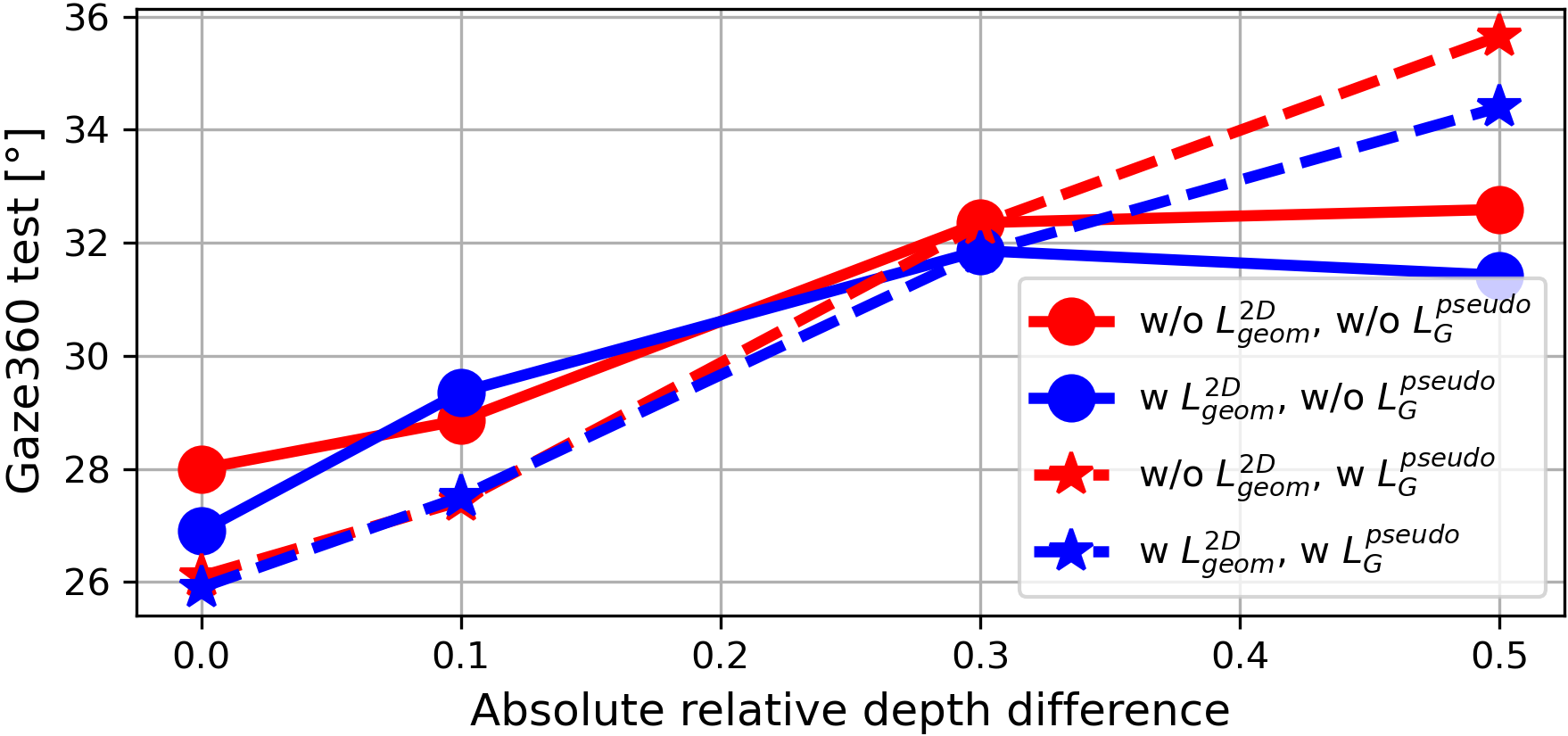}
    \caption{Purely weakly-supervised performance of CMU Panoptic on Gaze360, with added relative depth noise ($\mu=0$, $\sigma=\{0.1, 0.3, 0.5\}$), when trained with different combinations of LAEO losses ($L^{3D}_{geom}$ is always on). With the $L^{2D}_{geom}$ loss included, performance degrades more gracefully on increasing depth noise versus without. Plots show median values across 4 different training runs initialized with different network weights.}
     \label{fig:less_degrade}
\end{figure}

The 2D eye position on the image plane can be estimated without depth ambiguity and is more reliable than the 3D eye position. To quantify the contribution of $L^{2D}_{geom}$ to the overall performance of our system, we add increasing noise ($z$-only) as a ratio of the absolute ground truth depth of the 3D eye positions to subjects under LAEO in the CMU Panoptic dataset, train various purely weakly-supervised configurations (as described in Sec.~4.1 of the main paper) with and without $L^{2D}_{geom}$ and evaluate on Gaze360 (Fig.~\ref{fig:less_degrade}).
While we see gaze prediction accuracy deteriorate with increasing depth noise, the inclusion of $L^{2D}_{geom}$ constrains gaze ambiguity and reduces the degradation of gaze estimates. Besides this, we also observe that including $L^{2D}_{geom}$ makes gaze predictions more consistent and reduces the standard deviation of errors on Gaze360's test set (CMU Panoptic: 27.0$^\circ \rightarrow$ 23.7$^\circ$ and AVA-LAEO: 23.6$^\circ \rightarrow$ 19.8$^\circ$).

\section{Detailed Cross-dataset Performance}
\begin{figure}
    \centering
    \includegraphics[width=\linewidth]{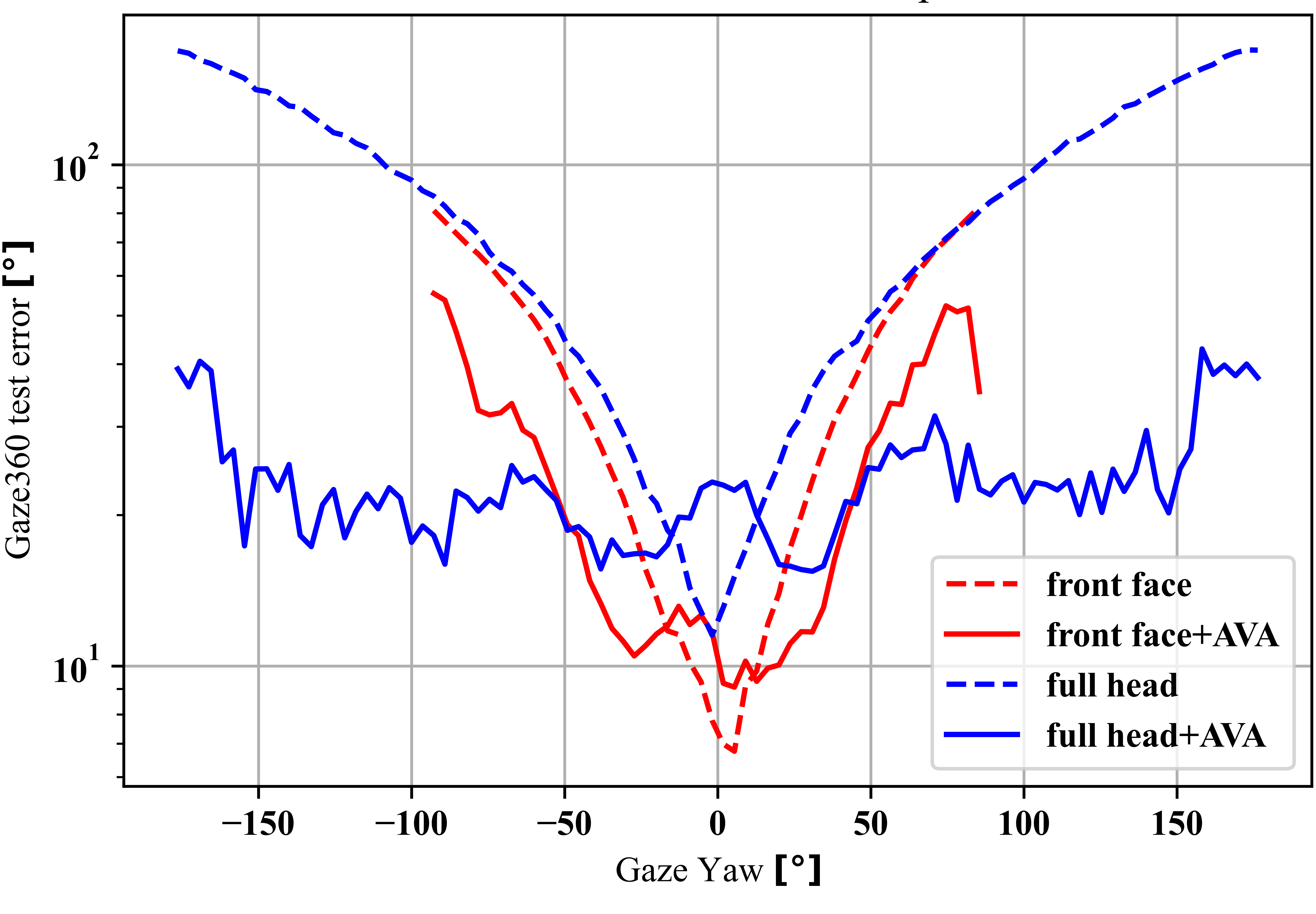}
    \caption[Caption for LOF]{Reduction in gaze error on the Gaze360 test set on jointly training with GazeCapture and AVA-LAEO. The dashed curves are for training with GazeCapture only and the solid ones are for jointly training with GazeCapture and AVA-LAEO. Each curve represents the mean of samples in bins 1.8$^\circ$ wide and the bins with 20 samples or less are discarded. The vertical axis is represented in $\mathrm{log}$ scale. Lower is better.}
    \label{fig:cross_gazecap}
\end{figure}

\begin{figure}
    \centering
    \includegraphics[width=\linewidth]{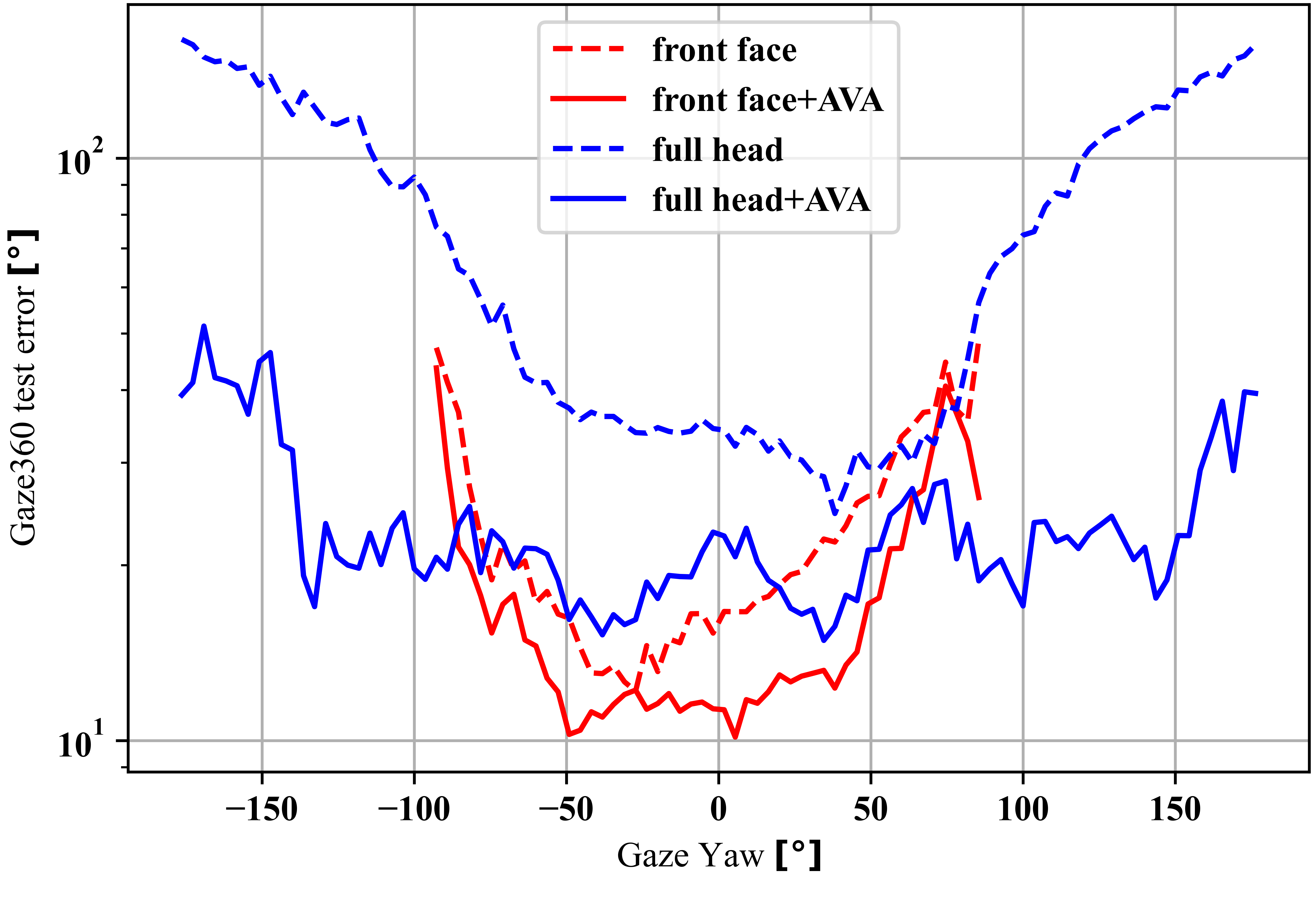}
    \caption[Caption for LOF]{Reduction in gaze error on the Gaze360 test set on jointly training with ETH-XGaze and AVA-LAEO. The dashed curves are for training with ETH-XGaze only and the solid ones are for jointly training with ETH-XGaze and AVA-LAEO. Each curve represents the mean of samples in bins 1.8$^\circ$ wide and the bins with 20 samples or less are discarded. The vertical axis is represented in $\mathrm{log}$ scale. Lower is better.}
    \label{fig:cross_xgaze}
\end{figure}

For the cross-dataset experiment described in Sec.~4.2 and Table~2 of the main paper, we additionally analyze the variation in gaze errors with varying gaze yaw angles on the Gaze360 test set. We consider the case of training with (a) GazeCapture only (dashed curves in Fig.~\ref{fig:cross_gazecap}) or (b) with GazeCapture and AVA-LAEO in (solid curves Fig.~\ref{fig:cross_gazecap}). The corresponding curves for training with (a) ETH-XGaze only or (b) with ETH-XGaze and AVA-LAEO are shown in Fig.~\ref{fig:cross_xgaze}. The blue curves show performance on the entire Gaze360 test set, while the red curves are for its subset containing frontal faces only.

The AVA-LAEO dataset exhibits a large distribution of extreme gaze angles as the LAEO activity largely consists of people with side profiles fixating at each other (see Fig.~1 and Fig.~2 in main paper and Fig.~\ref{fig:good_bad_ava} in the supplementary for examples). This conveniently augments datasets with narrow gaze distributions, \textit{e.g.}, GazeCapture (dashed versus solid curves in Fig.~\ref{fig:cross_gazecap}), which is largely concentrated about gaze pitch and yaw values of zero (from Fig.~3 of the main paper) and helps them generalize better to Gaze360. The AVA-LAEO dataset also contains a large appearance variability because of being collected from in-the-wild videos, which positively augments datasets collected indoors only, \textit{e.g.}, ETH-XGaze (dashed versus solid curves Fig.~\ref{fig:cross_xgaze}) and helps it generalize better to Gaze360 as well. On jointly training either the GazeCapture or ETH-XGaze dataset with AVA-LAEO, we see a significant boost in their performance on all head crops from Gaze360, including faces with large profile views (blue curves in Fig.~\ref{fig:cross_gazecap} and Fig.~\ref{fig:cross_xgaze}). Interestingly, adding the AVA-LAEO dataset improves cross-domain performance of GazeCapture and ETH-XGaze on Gaze360's frontal face crops as well (red curves in Fig.~\ref{fig:cross_gazecap} and Fig.~\ref{fig:cross_xgaze}).

\section{Data Pre-processing}
We first describe in detail how we pre-process the CMU Panoptic (\textit{haggling} activity subset) and the AVA-LAEO datasets. Then we analyze the effect of the simplifying assumptions that we employed to estimate scene geometry (as described in Sec.~3.3 of the main paper) on the reliability of 3D gaze annotations derived from real-world LAEO data.

\subsection{CMU Panoptic}
The CMU Panoptic dataset contains 31 views captured from high-definition cameras within a dome with available accurate body/facial 3D landmark locations and camera intrinsic and extrinsic parameters. This enables us to compute each subject's head position and orientation with respect to any scene camera. Such a convenient setup allows us to quickly gather our own large-scale gaze dataset by leveraging the LAEO constraint. However, this dataset does not contain explicit information about the presence or absence of the LAEO activity in video frames. So we use a semi-automatic procedure to label the video frames in it with LAEO activity labels. We use the pre-trained Gaze360 static network~\cite{Kellnhofer2019} to estimate gaze for every subject from multiple frontal views (\textit{i.e.}, if a given face is oriented within $\pm90^\circ$ of a camera's principal axis). These gaze estimates are then transformed to world co-ordinates and their pair-wise cosine distance is computed between every subject pair present in a frame. A pair of gaze vectors for two subjects are assumed to be under LAEO when their angular separation (with one of the vectors being inverted) from each other and the 3D line joining their cyclopean 3D eyes is $<20^\circ$. A pair of subjects is treated to be in LAEO when at least 4 of its gaze pairs from multiple views are classified as being in LAEO. The nature of the \textit{haggling} activity ensures that only a single pair may ever exhibit LAEO. Frames with none or multiple LAEO pair detections are removed from the analysis.

\begin{figure}
    \centering
    \includegraphics[width=\linewidth]{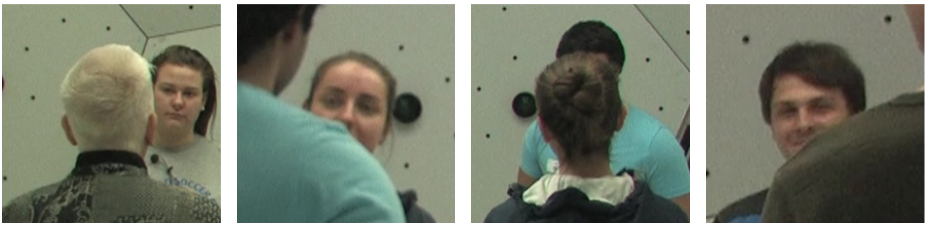}
    \caption{Examples of discarded CMU Panoptic frames from our experiments.}
    \label{fig:discarded_pano}
\end{figure}

We experience two corner cases: a) facial features of certain subjects can be blocked from view by another subject in the scene and b) multiple subjects may appear within the same head bounding box (see Fig.~\ref{fig:discarded_pano}). To mitigate this issue, we first compute a facial bounding box surrounding a subject's ears, eyes and nose keypoints. Next, we compute a bounding box around every subject's body. Views with facial bounding boxes overlapping with body bounding boxes of other subjects (\textit{i.e.}, with a bounding box IOU score $\geq$0.01) are discarded from the analysis. This in-turn results in missing gaze values in the central $\pm 15^\circ$ gaze pitch and yaw distribution region (see Fig.~3 in the main paper).

\begin{figure}[t]
    \centering
    \includegraphics[width=\linewidth]{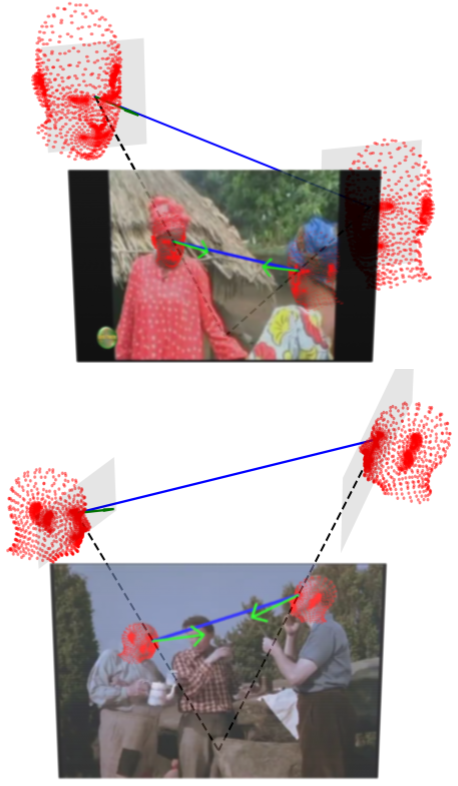}
    \caption{(\textbf{Top}) a positive and (\textbf{bottom}) a negative example of scene geometry reconstruction from the AVE-LAEO dataset. Notice the incorrect 3D head placement for the right most subject in the bottom example with respect to $z$ depth. The subject on the right is clearly closer to the camera (in terms of $z$ depth) than the subject on the left, but is incorrectly estimated as being further from it. This, in turn, results in noisy 3D gaze labels.}
    \label{fig:good_bad_ava}
\end{figure}

\subsection{AVA-LAEO}
The availability of 3D head poses and landmarks is a vital requirement for computing our LAEO losses. These annotation, however are not available in the AVA-LAEO dataset. We utilize dense 2D-3D correspondence predictions derived from DensePose~\cite{Guler2018} to fit the SMPL 3D head model to every detected subject within a LAEO annotated frame from the AVA training set~\cite{Gu2018} with LAEO annotations provided by Marin-Jimenez~\etal~\cite{Marin-Jimenez2019}. To improve computational efficiency while deriving these correspondences, we utilize up to 1,000 2D pixels detected by DensePose, which belong to a subject's head. To ensure that every detected facial region is well represented while computing 3D head pose, we uniformly sample 2D pixels based on their distance from the mean 2D head location. However, incorrect head-pose estimates due to incorrect 2D-3D correspondences are inevitable. See Fig.~\ref{fig:good_bad_ava} for a positive and a negative example of head pose fitting, where the latter results in noisy gaze labels for the AVA-LAEO dataset.

\subsection{Reliability of LAEO 3D Gaze Labels}
When scene geometry is unknown (\textit{e.g.}, in real-world LAEO datasets), 3D gaze labels derived from LAEO are indeed noisy. We introduce various constraints while training our system to counter this issue, and show results on both controlled (CMU Panoptic) and in-the-wild (AVA-LAEO) datasets. Yet, as a rough estimate, we compare the angular separation between 3D gaze derived from the approximate scene geometry (described in Sec.~3.3 of the main paper) and its ground truth values using a subset of 3495 images from the CMU Panoptic dataset. On average, we observe a $14.8^\circ$ gaze label error and an absolute relative depth difference of 0.3 between the ground truth and estimated subject depths when both 2D cyclopean eye points and the subjects' $z$ depths are estimated, and the focal length is assumed to be the largest image dimension. Replacing with accurate focal length reduces gaze label error to $10.1^\circ$ and using accurate 2D cyclopean eye centers further reduces it to $8.84^\circ$. Additionally, the assumption that people look at each others' 3D eye centers introduces $<4.3^\circ$ gaze error for subjects located $>500$mm apart. These label errors are significantly smaller than those encountered in cross-dataset ($\sim30^\circ$ from~\cite{Zhang2020ETHXGaze}) and semi-supervised ($>25^\circ$ from Fig.~4 of the main paper) training for Gaze360 making LAEO data a reliable source of supervision for 3D gaze learning in physically unconstrained settings.

\end{document}